\documentclass[10pt,twocolumn,letterpaper]{article}

\usepackage[pagenumbers]{wacv} 

\usepackage{booktabs}
\usepackage{threeparttable}
\usepackage{xcolor}
\usepackage{amsfonts}
\usepackage{multirow}
\usepackage{makecell}
\usepackage{amssymb}
\usepackage{amsthm}
\usepackage{comment}

\definecolor{wacvblue}{rgb}{0.21,0.49,0.74}
\usepackage[pagebackref,breaklinks,colorlinks,allcolors=wacvblue]{hyperref}

\def\wacvPaperID{297} 
\def\confName{WACV}
\def\confYear{2027}

\title{X-Splat: Gaussian Splatting for 3D CBCT Generation \\ from Single Panoramic Radiograph}

\author{
Tomasz Szczepański\textsuperscript{1,}\thanks{\texttt{t.szczepanski@sanoscience.org}} \quad
Szymon Płotka\textsuperscript{2} \quad
Michal K. Grzeszczyk\textsuperscript{1} \quad \\
Tomasz Trzciński\textsuperscript{3,4} \quad
Arkadiusz Sitek\textsuperscript{5} \\[.5em]
\textsuperscript{1}Sano Centre for Computational Medicine, Poland \quad
\textsuperscript{2}Jagiellonian University, Poland \quad \\
\textsuperscript{3}Warsaw University of Technology, Poland \quad
\textsuperscript{4}Research Institute IDEAS, Poland \quad \\
\textsuperscript{5}Harvard Medical School, USA
}

\begin{document}
\maketitle
Generating a 3D dental volume from a single panoramic radiograph (PXR) could provide a low-radiation alternative to Cone-Beam Computed Tomography (CBCT), but the problem is highly underdetermined: panoramic acquisition integrates 3D attenuation along curved X-ray paths into a 2D image, leaving depth-resolved anatomy unobserved. Existing implicit and generative approaches often produce oversmoothed geometry or anatomically inconsistent hallucinations, lacking geometry-driven supervision and relying on smooth representations unable to precisely localize sharp anatomical boundaries. We propose X-Splat, the first Gaussian Splatting framework for generating CBCT-like 3D dental volumes from a single PXR. X-Splat uses the known panoramic acquisition geometry as a generation scaffold: learnable anisotropic Gaussian primitives are initialized along the X-ray paths that formed the input image and adjusted in a single feed-forward pass, constrained by Beer--Lambert reprojection and multi-view radiographic training supervision. A lightweight residual refiner adds dataset-level anatomical priors without overriding the geometry already resolved by the Gaussians. We train on synthetic PXR--CBCT pairs, enabling direct volumetric supervision without paired real scans. We further introduce segmentation-based geometry-aware metrics, providing the first evaluation of PXR-based generation over maxillofacial anatomy. X-Splat outperforms NeRF- and GAN-based baselines, recovering individual teeth, cortical boundaries, and alveolar structure, including the mandibular canal which prior methods fail to reconstruct. Code will be available at \href{https://github.com/tomek1911/X-Splat}{\texttt{github.com/X-Splat}}.    
\section{Introduction} \label{sec:intro}

\begin{figure*}[t]
\centering
\includegraphics[width=\textwidth]{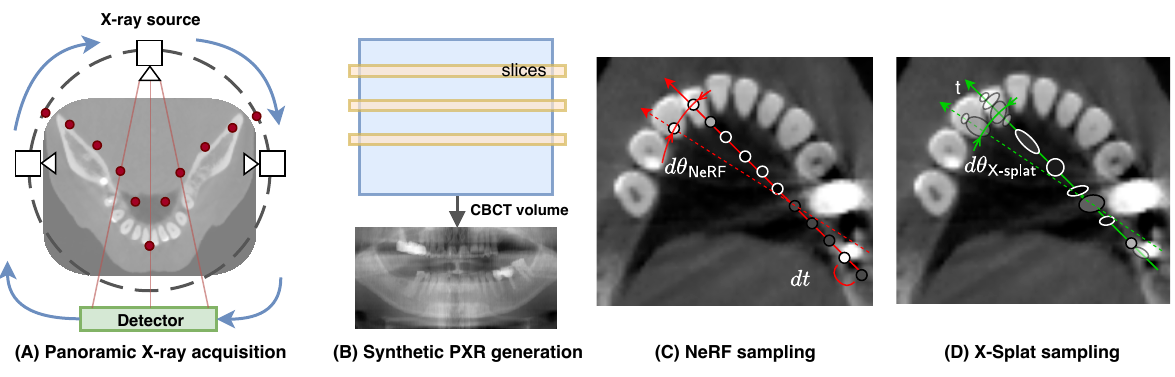}
\caption{(A) A panoramic X-ray (PXR) is acquired by rotating an X-ray source and detector around the patient's head, producing a single 2D projection encoding the full dental anatomy. (B) Synthetic PXRs are generated from a CBCT volume by casting rays through axial slices and accumulating attenuation according to Beer-Lambert law. (C) NeRF-based generation samples points at fixed intervals along each ray $dt$, unable to adapt to anatomical structure and leaving gaps between sparse rays unresolved. (D) \textbf{X-Splat} 3D Gaussians are anisotropic, learnable and parameterized along each ray $t$. They dynamically travel from their anchors, rotate and elongate to align with anatomical structures, efficiently filling inter-ray space.
}
\label{fig:intro_fig}
\end{figure*}

Inferring 3D anatomy from 2D projections is a fundamental problem across imaging science, physics, and computer vision, with particular clinical importance in dentistry, where Cone-Beam Computed Tomography (CBCT) remains the standard for 3D maxillofacial imaging by acquiring hundreds of 2D projections over a full arc to reconstruct a sub-millimetre attenuation volume~\cite{schulze2020cone}. Its diagnostic superiority over 2D imaging is well established, from detecting cortical destruction and root resorption to depicting the relationship between molar roots and the mandibular canal~\cite{y2017differences}. Yet CBCT is acquired only when clinically indicated, whereas panoramic radiography (PXR) is widely available, low-cost, and substantially lower in radiation dose~\cite{rozylo2020panoramic, pauwels2012effective}. Consequently, routine dental care often produces longitudinal archives of panoramic images, while CBCT data remain comparatively sparse. Inferring a CBCT-like volume from a single PXR would extend 3D dental analysis to large retrospective panoramic-image archives and to settings where CBCT is unavailable~\cite{macdonald2023cone, minhas2024artificial}. However, the problem is a highly underdetermined inverse problem: a panoramic projection integrates 3D attenuation along rays into  a single 2D image, leaving depth-resolved anatomy unobserved (Fig.\ref{fig:intro_fig}). Unlike sparse-view CT, where multiple projections constrain depth~\cite{zha2022naf, liu2024geometry, zang2021intratomo}, generating a 3D volume from a single-view requires strong anatomical and geometric priors as necessary constraints.\\
\indent
Existing PXR-to-3D methods~\cite{song2021oral, park2024nebla, li20243dpx, liang2020x2teeth, park2023occudent, ma2024px2tooth} leave three gaps that motivate our work. First, they rely on convolutional encoder-decoders or implicit neural
fields that represent the volume as a dense grid or continuous scalar
field; such representations are smooth by construction and
systematically under-resolve the sharp interfaces, such as root
boundaries, cortical bone surfaces, and the mandibular canal cortex,
that carry the most diagnostic value. Implicit paradigms like NeRF~\cite{mildenhall2021nerf} sample at fixed intervals along each ray and cannot adapt to anatomical structure, while GAN-\cite{ying2019x2ct,goodfellow2014generative} and diffusion-based~\cite{ho2020denoising, wang20253d} generators risk anatomically inconsistent hallucinations and often remain computationally prohibitive at CBCT resolution. Second, prior methods either require additional input beyond the PXR,
such as dental arch curves~\cite{song2021oral}, or produce volumes
from a single image that are locally plausible but globally inconsistent
with real dental topology. Third, evaluation remains dominated by intensity-based metrics such as SSIM and PSNR, sometimes with coarse global Dice Score~\cite{park2024nebla}; recent sparse-view studies show these aggregate voxel errors without reflecting whether clinically relevant structures are correctly recovered, so substantial structural inaccuracy can pass unnoticed~\cite{lin2026pixel}. A detailed review of related work is provided in Appendix \ref{app:related}.\\
\indent
To address these limitations, we propose \textbf{X-Splat}, a Gaussian Splatting framework for single-view 3D CBCT-like volumes from a panoramic radiograph. We formulate single-PXR 3D dental volume estimation as \emph{geometry-constrained generation}: unlike tomographic reconstruction, which resolves depth from multiple projections, a single PXR leaves the volume underdetermined, requiring conditional generation under strong geometric priors. Rather than a smooth implicit field, we represent the volume with localized anisotropic 3D Gaussian primitives~\cite{kerbl20233d} whose explicit geometry naturally encodes sharp boundaries and can be optimized via physics-consistent radiative rendering aligned with the Beer–Lambert attenuation model~\cite{cai2024radiative, zha2024r2gaussian}. Gaussians are initialized along panoramic acquisition rays and predicted in a feed-forward pass, then displaced, rotated, and anisotropically scaled to follow anatomical boundaries and fill the space between sparsely separated rays (Fig.\ref{fig:intro_fig}D). Finally, we evaluate generated volumes' quality using a ToothFairy3-trained segmentation network~\cite{bolelli2024segmenting,bolelli2025segmenting,
bolelli2026multi,szczepanski2025gepar3d,szczepanski2026morphology} to
derive geometry-aware metrics over individual teeth, the mandibular canal
(MC), and the five major maxillofacial structures, providing a more
faithful assessment of structural fidelity than intensity measures alone. Our main contributions are as follows:
\begin{enumerate}

\item We propose \textbf{X-Splat}, the first Gaussian Splatting framework for generating CBCT-like 3D dental volumes from a single panoramic radiograph. Explicit anisotropic Gaussian primitives, constrained by Beer--Lambert reprojection and multi-view radiographic supervision, provide a geometry-grounded alternative to the smooth implicit fields and dense voxel decoders of prior PXR-to-3D methods.

\item From a single PXR alone, X-Splat recovers fine dental and maxillofacial structures that existing methods systematically miss, including individual tooth roots, cortical boundaries, and the mandibular canal. This is enabled by a ray-anchored Gaussian representation in which a shared feed-forward network predicts each primitive's displacement, orientation, scale, and density using the known panoramic acquisition geometry as an acquisition scaffold.

\item We combine geometry-driven Gaussian generation with a lightweight residual refiner constrained by multi-view DRR supervision, adding dataset-level anatomical detail while keeping the generated volume tied to the input projection. We further introduce geometry-aware structural metrics beyond conventional image-quality measures, evaluating recovery of individual teeth, cortical boundaries, alveolar structure, and the mandibular canal.

\end{enumerate}
\section{Methods}%
\label{sec:methods}

\begin{figure*}[t]
\centering
\includegraphics[width=\textwidth]{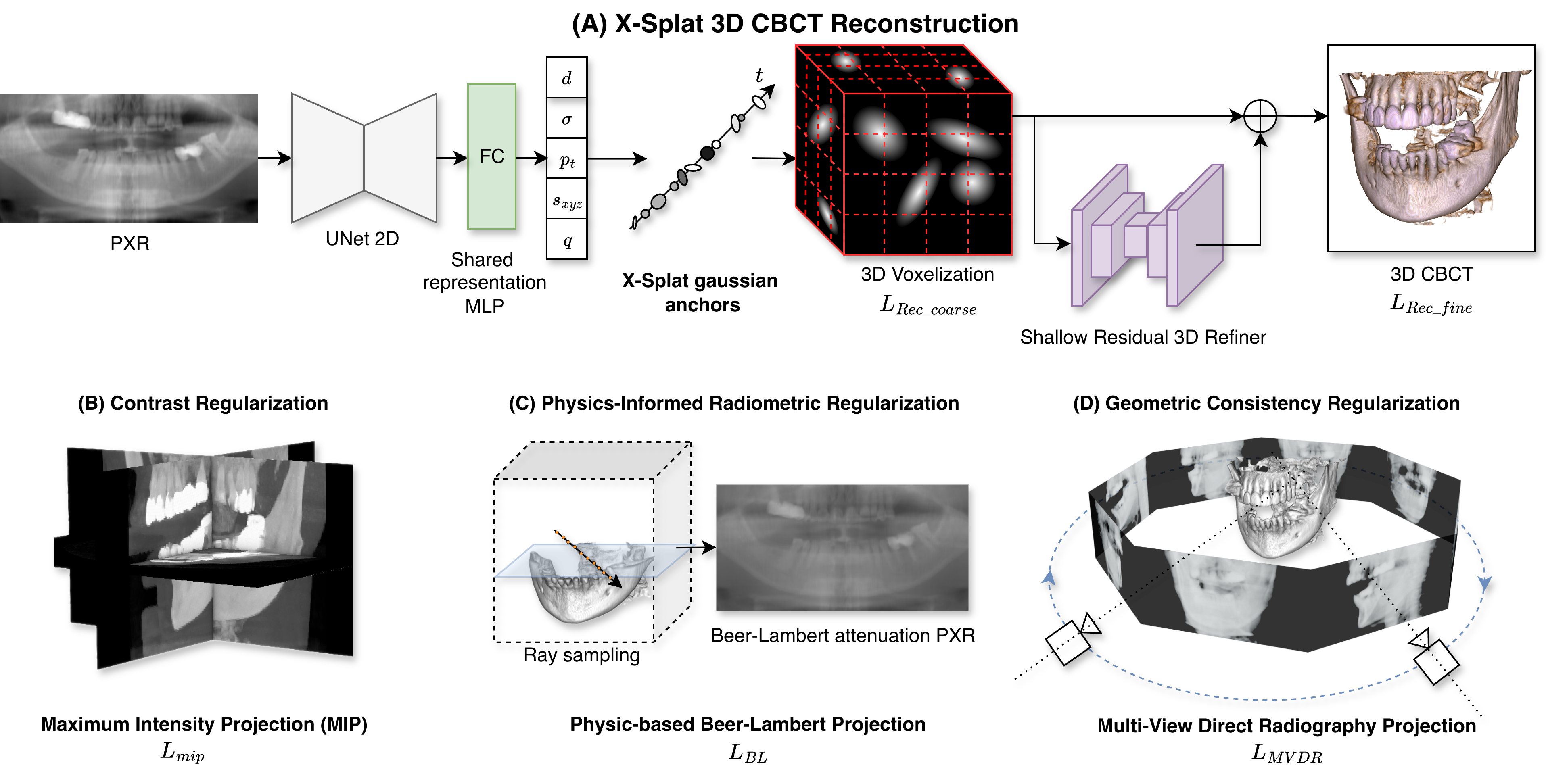}
\caption{\textbf{Method overview and regularization scheme}. \textbf{(A)} A single panoramic X-ray (PXR) is processed by a 2D U-Net encoder and shared MLP to predict a set of 3D Gaussians, which are voxelized into a coarse CBCT volume. A residual 3D U-Net refiner then produces the final volume. Both the coarse and refined volumes are supervised with three regularizations: \textbf{(B)} Contrast Regularization (MIP-based), \textbf{(C)} Radiometric Regularization (Beer-Lambert law), and \textbf{(D)} Geometric Regularization (multi-view ray consistency). The refined volume is additionally supervised via perceptual loss on its 2D projections.
}
\label{fig:overview}
\end{figure*}


We design X-Splat using three complementary choices. \textbf{(i)~Geometry-anchored initialization:} we exploit the known panoramic acquisition geometry, anchoring 3D Gaussian primitives along the predefined ray trajectories that map 2D pixels to 3D space, concentrating capacity where the input is informative without per-scene optimization. \textbf{(ii)~Feed-forward prediction and refinement:} a 2D U-Net encoder and a single MLP, shared across all Gaussian attributes and anchors, predict all Gaussian parameters from the input and each anchor's positional encoding; the Gaussians are rasterized into a coarse volume, which a lightweight 3D residual U-Net refines, adding population-level anatomical priors as a small correction that keeps the output tethered to the input. \textbf{(iii)~Multi-view consistency:} during training, additional azimuthal projections supervise the volume, contracting the feasible set toward anatomically consistent solutions. Figure~\ref{fig:overview} illustrates the pipeline; the following subsections detail each component.

\subsection{Problem Formulation}

Let $\mathbf{I} \in \mathbb{R}^{H_p \times W_p}$ denote a simulated panoramic
X-ray (SimPX) rendered from a ground-truth CBCT volume
$V^* \in \mathbb{R}^{H \times W \times D}$ using the panoramic ray geometry
and effective Beer--Lambert rendering convention of NeBLa~\cite{park2024nebla}.
This provides a calibrated computational forward model of panoramic image
formation in which scanner-dependent spectral and intensity effects are
absorbed into effective attenuation and rendering parameters. For a ray
$\mathbf{r}(t)$, the projected intensity is:

\begin{equation}
  \mathcal{F}(V,\theta)(h,w)
    = \frac{1 - I_0\,\exp\!\Bigl(-\beta\sum_k \mu_k\,\Delta s\Bigr)}{p_{\max}},
  \label{eq:beer_lambert}
\end{equation}

\noindent where $\mu_k$ denotes the effective attenuation at sample $k$ along
the ray, $\Delta s$ is the voxel step size, and $I_0$, $\beta$, and
$p_{\max}$ are calibrated rendering constants. The inverse problem is to
recover $\hat{V}$ from a single observation
$\mathbf{I} = \mathcal{F}(V^*,\theta_0)$.%

\subsection{SimPX Ray Geometry}
\label{sec:geometry}

We adopt the SimPX synthesis pipeline and focal-trough ray geometry
from NeBLa~\cite{park2024nebla}. The panoramic focal trough is modelled by a quadratic curve
$f(x) = 0.01(x \pm 100)^2$, where $x$ denotes horizontal coordinate in the axial CBCT slice, that defines 21 rotation centers: (see Fig.\ref{fig:intro_fig} (A))
$\mathbf{c}_i$, $i=0,\ldots,20$, placed on the axial plane of a
$256\!\times\!256$ CBCT slice.
At each centre $\mathbf{c}_i$, rays are rotated by an angle $\theta_i$
that varies with the dental region (denser near the molars) to simulate
the variable angular sweep of a real panoramic scanner.
Along each ray, $K=200$ points are uniformly sampled within the volume
boundary and rendered to a single pixel via Eq.~\eqref{eq:beer_lambert}.
The resulting SimPX has $H_p$ rows (one per CBCT slice in $Z$) and $W_p$
columns equal to the total number of rays, producing a 2D image whose
pixels map to known 3D ray trajectories.
This explicit ray-pixel correspondence is the geometric prior that
X-Splat exploits for anchor placement, as described next.

\subsection{Gaussian Anchor Initialization}
\label{sec:anchors}

The ray geometry provides a uniquely informative spatial prior.
Every pixel $(h,w)$ of the SimPX corresponds to a known 3D ray
$\mathbf{r}_{h,w}(t)$, and the pixel intensity encodes the integrated
attenuation along that ray.
Rather than learning anchor positions from scratch through adaptive
densification, we exploit this prior directly:
we place one Gaussian anchor per ray sample, yielding
$N_g \approx 6\mathrm{M}$ anchors
$\{\mathbf{p}_j\}_{j=1}^{N_g}$ that densely cover the dental arch and
jaw region from the outset. This concentrates representational capacity where the input is informative, avoiding the costly exploration phase that unconstrained Gaussian methods require.

\subsection{3D Gaussian Splatting}
\label{sec:gaussians}

\noindent
\textbf{Shared MLP.} A single MLP $f_\phi$, shared across all $N_g$
anchors, predicts each Gaussian's displacement, scale, rotation, and
density in one forward pass. We extend the MLP conditioning scheme of
NeBLa~\cite{park2024nebla}, combining global image features with
per-point positional encoding, to the Gaussian Splatting domain.
Per-scene optimization methods solve these jointly through iteration
with adaptive densification and non-uniform learning rates per
attribute~\cite{kerbl20233d,zha2024r2gaussian}; our feed-forward
setting must resolve all parameters simultaneously and uniformly from
the input image alone. Sharing weights across anchors is key: rather
than predicting each attribute through separate heads, the network
builds joint inductive biases over the full anatomy, since tissue of
a given density at a given location implies a characteristic scale and
orientation, which would be underexploited with a multi-head design.
The MLP predicts:
 
\begin{equation}
  (\hat{\delta t}_j,\; \mathbf{s}_j,\; \mathbf{q}_j,\; \alpha_j)
    = f_\phi\!\left(\mathbf{F}(\mathbf{p}_j) + \mathrm{PE}(\mathbf{p}_j)\right),
  \label{eq:mlp}
\end{equation}
 
\noindent where $\mathbf{F}(\mathbf{p}_j)$ are UNet features read directly at the
pixel location on ray corresponding to anchor $\mathbf{p}_j$ in the SimPX,
$\mathrm{PE}(\mathbf{p}_j)$ is the positional encoding,
$\hat{\delta t}_j$ is a raw scalar displacement,
$\mathbf{s}_j\in\mathbb{R}^3$ the log-scale,
$\mathbf{q}_j\in\mathbb{R}^4$ a unit quaternion, and
$\alpha_j$ the density.

\noindent
\textbf{Constrained axial-plane movement.} Each Gaussian is permitted to move only along its anchor ray within the
axial (XY) plane.
This constraint is motivated by the structure of the SimPX: because
rays run perpendicular to the dental arch and the CBCT is sampled
slice-by-slice along $Z$, allowing inter-slice ($Z$-axis) Gaussian
migration would corrupt the dense axial coverage.
The scalar displacement is bounded by $\delta t_j = \tanh(\hat{\delta t}_j)\cdot\delta t_{\max}$, where $\delta t_{\max} = 32\;\text{vox}$, and projected onto a 3D offset via the ray's axial unit direction $\hat{\mathbf{d}}_{w_j}\in\mathbb{R}^2$:

\begin{equation}
  \Delta\mathbf{p}_j =
  \Bigl[\delta t_j\cdot\frac{\hat{\mathbf{d}}_{w_j}}{\mathbf{h}_{XY}},\;0\Bigr],
  \label{eq:delta_xyz}
\end{equation}

\noindent where $\mathbf{h}_{XY}=[H/2,\,W/2]$ maps voxel displacement
to the $[-1,1]^3$ coordinate system and the zero $Z$-component enforces
the no-inter-slice constraint.
The final Gaussian centre is
$\tilde{\mathbf{p}}_j = \mathbf{p}_j + \Delta\mathbf{p}_j$.

\noindent
\textbf{Axial-plane rotation.} Gaussian rotation is restricted to the axial plane (rotation around the
superior-inferior $Z$-axis), again to prevent cross-slice interference
and to keep each Gaussian's spatial footprint aligned with its ray.

\noindent
\textbf{Scale initialization and clamping.} All scale axes are initialized to $\sigma_0 = 0.25\;\text{px}$, close to the inter-ray spacing around the dental arch, and clamped during training to $s_{\max} = 1.0\;\text{px}$. At this ceiling, $95\%$ of each Gaussian's density is contained within a radius of $2\sigma = 2.0\;\text{px}$, matching the observed inter-ray distance near the teeth and ensuring Gaussians do not grow large enough to bridge anatomically distinct structures on neighbouring rays. Starting small and allowing growth lets the model first place Gaussians
accurately and then expand them to fill inter-ray gaps, avoiding the
degenerate all-diffuse initialization that large initial scales would produce.

\noindent
\textbf{Coarse volume synthesis.} The $N_g$ Gaussians are rasterized into a coarse volume $\hat{V}_c\in\mathbb{R}^{H\times W\times D}$ via the differentiable
R2-Gaussian voxelizer~\cite{zha2024r2gaussian}, which corrects an integration bias in standard 3DGS for X-ray physics:

\begin{equation}
  \hat{V}_c(\mathbf{x}) =
  \sum_j \alpha_j\,
  \exp\!\left(
    -\tfrac{1}{2}(\mathbf{x}-\tilde{\mathbf{p}}_j)^\top
    \boldsymbol{\Sigma}_j^{-1}
    (\mathbf{x}-\tilde{\mathbf{p}}_j)
  \right),
  \label{eq:voxelizer}
\end{equation}

\noindent where $\boldsymbol{\Sigma}_j=\mathbf{R}_j\,\mathrm{diag}(\mathbf{s}_j^2)\,\mathbf{R}_j^\top$
and $\mathbf{R}_j$ is the rotation matrix derived from $\mathbf{q}_j$.
Gradients flow through the voxelizer back to $f_\phi$, enabling
end-to-end training.

\subsection{Residual Refiner}

The Gaussian coarse volume $\hat{V}_c$ already carries the geometric
structure of the generation: because all projection losses are applied
directly to $\hat{V}_c$, the MLP is explicitly trained to solve the 3D
lifting problem from the 2D conditioning signal.
We additionally supervise $\hat{V}_f$, giving the refiner the role of
incorporating population-level anatomical priors (typical jaw morphology,
tooth arrangement, density distributions) learned at the dataset level,
while the geometric fidelity to the input PXR is anchored in the coarse
stage.\\
\indent
To enforce this division of responsibility we parameterize the refiner as
a residual correction.
A deliberately small 3D U-Net $r_\psi$ (see Appendix~\ref{app:architecture}) predicts:
 
\begin{equation}
  \hat{V}_f = \hat{V}_c + r_\psi(\hat{V}_c).
  \label{eq:refiner}
\end{equation}
 
The refiner is intentionally kept small (3 stages, max 128 feature
channels), in contrast to NeBLa's\cite{park2024nebla} refiner with 5 stages and a 1024
channel bottleneck, which risks memorizing population-level anatomy
over geometric fidelity. latexA large network trained on a fixed cohort risks memorizing
population-level anatomy regardless of the input, prioritizing dataset
statistics over the geometric constraints in the coarse volume.
By limiting capacity and restricting to a residual, the network can
only make small adjustments and cannot deviate substantially from the
geometrically grounded Gaussian prediction, favouring generalization
over memorization.The final convolutional layer is zero-initialized so that $r_\psi$
begins as a zero correction, ensuring $\hat{V}_f = \hat{V}_c$ early
in training and gradient signal flows through the Gaussian parameters
and MLP before the refiner contributes.

\subsection{Training Objective}

X-Splat is trained end-to-end with a composite loss applied independently
to both $\hat{V}_c$ and $\hat{V}_f$ with equal stage weights.
Losses are grouped into three categories calibrated after training
stabilized at approximately 100 epochs: geometric projection losses
($\approx70\%$), perceptual losses ($\approx15\%$), and volumetric
intensity loss ($\approx15\%$).

\noindent
\textbf{Volumetric intensity loss.} MSE between prediction and ground-truth CBCT penalizes global intensity
errors:
\begin{equation}
  \mathcal{L}_{\mathrm{rec}} = \|\hat{V} - V^*\|^2.
  \label{eq:rec_loss}
\end{equation}

\noindent
\textbf{Geometric projection losses.} Three complementary losses constrain the reconstructed 3D geometry.
The Beer-Lambert projection loss renders $\hat{V}$ back to a SimPX and
compares it to the input:
\begin{equation}
  \mathcal{L}_{\mathrm{proj}} =
    \|\mathcal{F}(\hat{V},\theta_0) - \mathbf{I}\|^2.
  \label{eq:proj_loss}
\end{equation}

Maximum-intensity projection (MIP) losses along all three anatomical axes
provide orthogonal structural supervision without requiring full volumetric
alignment:
\begin{equation}
  \mathcal{L}_{\mathrm{mip}} =
    \sum_{a\in\{x,y,z\}}
    \bigl\|\mathrm{MIP}_a(\hat{V}) - \mathrm{MIP}_a(V^*)\bigr\|^2.
  \label{eq:mip_loss}
\end{equation}

The multi-view DRR loss $\mathcal{L}_{\mathrm{mv}}$
(Eq.~\eqref{eq:mv_loss}, Section~\ref{sec:mv})
resolves front-to-back depth ambiguity by enforcing consistency across azimuthal projections.

\noindent
\textbf{Perceptual losses.} Following NeBLa~\cite{park2024nebla}, VGG-based perceptual losses are
applied to 2D projections of $\hat{V}_f$ to encourage high-frequency
anatomical detail.
Perceptual supervision covers the Beer-Lambert projection, all three MIP
projections, and eight representative multi-view DRR angles at
$\theta\!\in\!\{{\pm}37.5^{\circ},{\pm}52.5^{\circ},{\pm}67.5^{\circ},{\pm}82.5^{\circ}\}$
selected for symmetric angular coverage:
\begin{equation}
  \mathcal{L}_{\mathrm{perc}} =
    \sum_{P\in\mathcal{P}}
    \bigl\|\phi_{\mathrm{VGG}}(P(\hat{V}_f))
         - \phi_{\mathrm{VGG}}(P(V^*))\bigr\|^2,
  \label{eq:perc_loss}
\end{equation}
where $\mathcal{P}$ collects all supervised projection operators and
$\phi_{\mathrm{VGG}}$ extracts intermediate VGG feature maps.

\noindent
\textbf{Total loss.} The composite objective, 
\begin{align}
  \mathcal{L} =
    \underbrace{
      \lambda_{\mathrm{rec}}\,\mathcal{L}_{\mathrm{rec}}
    }_{\approx15\%}
  &+\underbrace{
      \lambda_{\mathrm{proj}}\,\mathcal{L}_{\mathrm{proj}}
     +\lambda_{\mathrm{mip}}\,\mathcal{L}_{\mathrm{mip}}
     +\lambda_{\mathrm{mv}}\,\mathcal{L}_{\mathrm{mv}}
    }_{\approx70\%} + \nonumber\\
  &+  \underbrace{
      \lambda_{\mathrm{perc}}\,\mathcal{L}_{\mathrm{perc}}
    }_{\approx15\%},
  \label{eq:total_loss}
\end{align}
is summed over both $\hat{V}_c$ and $\hat{V}_f$ with equal stage weights.
Loss weights were tuned once on a held-out validation case after training
had stabilized and were kept fixed thereafter.

\begin{table*}[t]
\centering
\begin{threeparttable}
\caption{Quantitative comparison of 3D CBCT generation from a single synthesized panoramic X-ray.
Intensity-based metrics (PSNR, SSIM, LPIPS) measure overall volume fidelity.
Geometry-aware metrics assess spatial generation accuracy independently of absolute
patient positioning:
\textbf{BA-ASD} (Big Anatomy Average Surface Distance, mm\,$\downarrow$) measures surface
accuracy over five major maxillofacial structures~\cite{bolelli2025segmenting};
\textbf{TVR} (Teeth Volume Recall, \%\,$\uparrow$) measures the fraction of ground-truth tooth
volume reconstructed, averaged per tooth class present in the patient;
\textbf{CVR} (Canal Volume Recall, \%\,$\uparrow$) measures the fraction of ground-truth
mandibular canal (MC) volume reconstructed;
\textbf{HV} (Hallucinated Volume, \%\,$\downarrow$) is the total volume of falsely predicted
tooth structures as a fraction of the ground-truth teeth volume.
Best per column in \textbf{bold}; second best \underline{underlined}. $\dagger$ - intensity metrics calculated only for arch offset sub-volume - see Fig. \ref{fig:qualitativeresults}, $\ddagger$ - uses dental arch geometry derived from GT segmentation labels (oracle geometry).}     
\label{tab:tab1}
\begin{tabular}{lccccccc}
\toprule
& \multicolumn{3}{c}{Intensity-based} & \multicolumn{4}{c}{Geometry-aware} \\
\cmidrule(r){2-4} \cmidrule(l){5-8}
Method & PSNR $\uparrow$ & SSIM $\uparrow$ & LPIPS $\downarrow$
       & BA-ASD $\downarrow$ & TVR $\uparrow$ & CVR $\uparrow$ & HV $\downarrow$ \\
\midrule
X2CT-GAN~\cite{ying2019x2ct} & $19.52_{1.00}$ & $66.33_{3.15}$ & $0.373_{0.029}$ & $58.60_{15.98}$ & $25.95_{23.86}$ & $1.88_{2.26}$ & $4.38_{8.00}$ \\
R2N2~\cite{choy20163d} & $23.11_{0.70}$ & $61.63_{1.28}$ & $0.324_{0.011}$ & $28.98_{19.01}$ & $36.08_{7.64}$ & $0.01_{0.02}$ & $19.27_{16.46}$ \\
Residual CNN~\cite{henzler2018single}  & $21.75_{1.02}$ & $\underline{76.00_{2.37}}$ & $\underline{0.286_{0.021}}$ & $13.06_{12.29}$ & $66.80_{9.58}$ & $\underline{27.62_{21.28}}$ & $14.95_{18.25}$ \\
GAN~\cite{goodfellow2020generative}    & $\underline{23.12_{0.67}}$ & $67.59_{1.96}$ & $0.297_{0.010}$ & $\underline{10.60_{16.59}}$ & $70.29_{10.50}$ & $9.11_{7.70}$ & $14.36_{15.04}$ \\
NeBLa~\cite{park2024nebla}  & $20.79_{0.45}$ & $74.42_{0.84}$ & $0.297_{0.006}$ & $35.67_{21.97}$ & $\underline{74.12_{3.64}}$ & $1.14_{1.27}$ & $\underline{3.99_{5.33}}$ \\
\cmidrule(l){1-1} \cmidrule(l){2-4} \cmidrule(l){5-8}
Oral-3D~\cite{song2021oral}   & $25.93_{0.44}$$^\dagger$ & $86.57_{1.23}$$^\dagger$ & $0.162_{0.019}$$^\dagger$ & $47.55_{18.27}$$^\ddagger$ & $76.58_{10.79}$$^\ddagger$ & $48.35_{22.73}$$^\ddagger$ & $1.07_{1.47}$$^\ddagger$ \\
\midrule
\textbf{X-Splat (Ours)}                & $\mathbf{23.27_{0.91}}$ & $\mathbf{79.62_{1.76}}$ & $\mathbf{0.259_{0.015}}$ & $\mathbf{2.39_{0.74}}$ & $\mathbf{84.96_{4.51}}$ & $\mathbf{67.33_{19.83}}$ & $\mathbf{2.96_{5.16}}$ \\
\bottomrule
\end{tabular}
\end{threeparttable}
\end{table*}

\subsection{Multi-view DRR Consistency Loss}
\label{sec:mv}
Reconstructing a 3D volume from a single 2D projection is fundamentally
ill-posed: the feasibility set
$\mathcal{S}{=}\{V:\mathcal{F}(V,\theta_0){=}I_{\theta_0}\}$
is infinite-dimensional, containing every configuration that preserves
Beer-Lambert integrals along $\theta_0$.
We resolve this with a training-time regularizer inspired by angular
coverage in CT~\cite{natterer2001mathematics} and sparse-view neural
rendering~\cite{mildenhall2021nerf,kerbl20233d}, supervising the predicted
volume against DRRs from $N{=}31$ azimuthal angles.
While a $180^{\circ}$ arc is theoretically sufficient for an ideal upright
patient~\cite{natterer2001mathematics}, dental CBCT patients exhibit
non-negligible pitch and roll, breaking projection symmetry and introducing
genuine parallax beyond $180^{\circ}$; we therefore use $\Delta\theta{=}7.5^{\circ}$
spacing over ${\pm}112.5^{\circ}$ ($225^{\circ}$ total), matching clinical half-scan
practice~\cite{scarfe2008cbct}. Unlike prior methods supervised only along anatomical axes, these Beer-Lambert DRRs expose azimuthal views unavailable from any other
loss, aimed at resolving depth and reducing geometric ambiguity. The loss penalizes any predicted volume $\hat{V}$ whose projections diverge from the GT DRRs:
\begin{equation}
  \mathcal{L}_{\mathrm{mv}} =
  \frac{1}{N}\sum_{i=1}^{N}
  \bigl\|\mathcal{F}(\hat{V},\theta_i) - \mathcal{F}(V^*,\theta_i)\bigr\|^2.
  \label{eq:mv_loss}
\end{equation}
Gradients flow through the differentiable renderer back into the Gaussian
parameters and MLP weights, penalizing configurations that satisfy the
single-view loss yet produce incorrect lateral projections. Under this
joint supervision the feasibility set
$\bigcap_{i=1}^{N}\{V:\mathcal{F}(V,\theta_i)=\mathcal{F}(V^*,\theta_i)\}$
contracts toward geometrically consistent generations, driving splats
toward true tissue boundaries.
\section{Experiments}%
\label{sec:experiments}

\begin{figure*}[t]
\centering
\includegraphics[width=\textwidth]{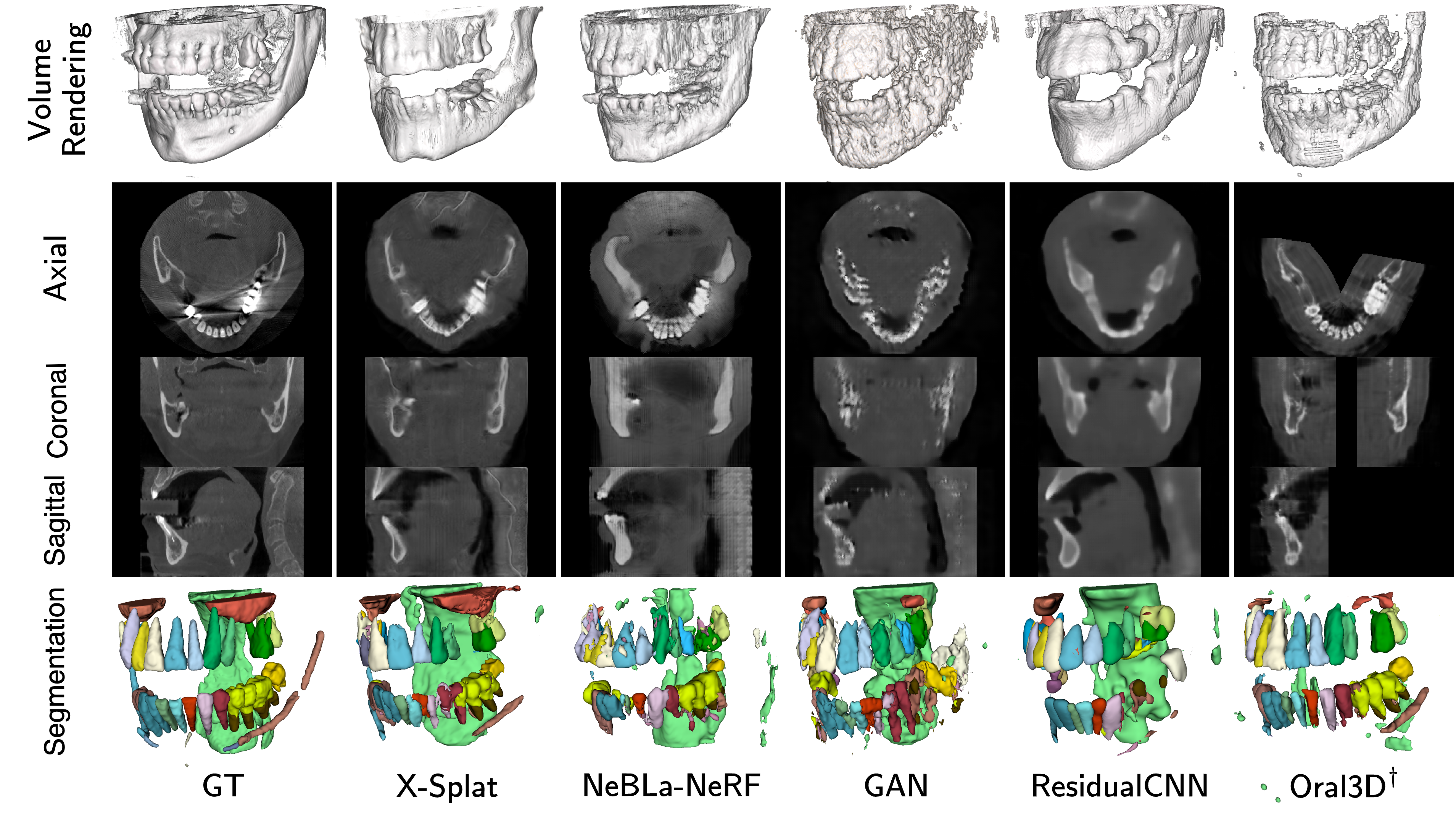}
\caption{Qualitative comparison of 3D CBCT generation from a single panoramic
X-ray. From top: volume rendering, axial, coronal, and sagittal
cross-sections, and 3D segmentation. X-Splat recovers cortical and
trabecular bone, maxillary sinuses, and the MC without hallucinating teeth
in edentulous regions; sharp cortical boundaries are visible as bright
shells in the cross-sections. Maxilla and mandible labels are hidden in
the segmentation row to reveal MC and sinus structures.
$^\dagger$Oral-3D uses ground-truth arch geometry and reconstructs only
the dental arch sub-volume; surrounding anatomy is absent.
}
\label{fig:qualitativeresults}
\end{figure*}

\noindent
\textbf{Dataset.} To train and evaluate, we use a subset of a publicly available ToothFairy3 \cite{2026MICCAI_toothfairy3,2024IEEEACCESS, 2025CVPR} dataset. We select scans with no more than 50\% missing dentition and a full maxillofacial field of view. Each scan includes the complete anatomy of all visible teeth, from crown to root apex, with no dental structures cropped or omitted. Applying these criteria to the full ToothFairy3 release yielded
60 eligible scans with paired segmentation labels. The dataset was randomly partitioned into training (n = 45), validation (n = 5), and test (n = 10) sets, while maintaining a comparable proportion of patients with missing teeth across all subsets.\\
\noindent
\textbf{Implementation details.} We train X-Splat exclusively on synthetic SimPX--CBCT pairs, simulating the panoramic projection from each training CBCT using Eq.~\eqref{eq:beer_lambert} and the scanning geometry of Section~\ref{sec:geometry}. This yields perfectly registered input--output pairs at no acquisition cost. Matched real PXR--CBCT pairs are exceedingly rare in clinical practice, as the two modalities are seldom acquired on the same day, and verifying sub-voxel registration between them is non-trivial. All implementation details can be found in \ref{app:implementation_details}.\\
\noindent
\textbf{Evaluation metrics.} We report intensity-based metrics (PSNR,
SSIM, LPIPS) and four geometry-aware metrics: Big Anatomy Average
Surface Distance (BA-ASD, mm$\,\downarrow$), Teeth Volume Recall
(TVR, \%$\,\uparrow$), Canal Volume Recall (CVR, \%$\,\uparrow$),
and Hallucinated Volume (HV, \%$\,\downarrow$). Full definitions
are in Appendix~\ref{app:evaluation_metrics}.\\
\noindent
\textbf{Comparison with state-of-the-art.}
We evaluate against six methods covering domain-specific PXR-to-3D
generation (NeBLa~\cite{park2024nebla}, Oral-3D~\cite{song2021oral}),
general volumetric prediction (R2N2~\cite{choy20163d}, Residual
CNN~\cite{henzler2018single}, GAN~\cite{goodfellow2014generative}), and
X-ray-to-CT synthesis (X2CT-GAN~\cite{ying2019x2ct}), all trained on the
same data for full craniofacial reconstruction. Oral-3D additionally
requires a dental arch curve (originally from an oral photograph); we
supply it from ground-truth labels, granting oracle access unavailable at
real inference. Full descriptions in Appendix~\ref{app:sota}.\\
\noindent
\textbf{Quantitative results.} Table~\ref{tab:tab1} reports results across all methods. X-Splat is best on every intensity metric (PSNR $23.27$\,dB, SSIM $79.62\%$, LPIPS $0.259$), but its decisive advantage lies in the geometry-aware metrics that better capture clinically relevant fidelity.\\
\indent
On surface accuracy, X-Splat attains a BA-ASD of $2.39$\,mm over the five major craniofacial structures, $4.4\times$ lower than the next oracle-free method (GAN, $10.60$\,mm). NeBLa~\cite{park2024nebla}, from which we inherit the ray geometry and projection supervision, reaches $35.67$\,mm, $14.9\times$ larger.Since both methods share the same panoramic geometry, MIP loss, and MSE loss, this gap reflects both the representational advantage of explicit Gaussians and the supervision strategies they enable: direct coarse-volume losses and multi-view DRR consistency, inapplicable to NeBLa's implicit field, leaving its large refiner under-constrained and prone to memorization over geometric fidelity.\\
\begin{table*}[t]
\centering
\caption{%
  Ablation study.
  \textit{MV}: multi-view DRR consistency loss;
  \textit{Crs}: supervision on the coarse Gaussian volume;
  \textit{Ref}: 3D residual U-Net refiner;
  \textit{Mov}: learned displacement of Gaussian anchors along rays.
  Geometry-aware metrics: BA-ASD - Big Anatomy Average Surface Distance\,mm\,$\downarrow$;
  TVR - Teeth Volume Recall\,\%\,$\uparrow$;
  HV — Hallucinated Volume\,\%\,$\downarrow$.
  Rows marked $^\dagger$ evaluate the coarse volume only (no refiner).
  Best per column in \textbf{bold}.
}
\label{tab:ablation}
\setlength{\tabcolsep}{2.8pt}
\begin{tabular}{l cccc | ccc | ccc}
\toprule
\multirow{2}{*}{Model} &
\multirow{2}{*}{MV} &
\multirow{2}{*}{Crs} &
\multirow{2}{*}{Ref} &
\multirow{2}{*}{Mov} &
\multicolumn{3}{c|}{Intensity-based} &
\multicolumn{3}{c}{Geometry-aware} \\
\cmidrule(lr){6-8}\cmidrule(lr){9-11}
& & & & & PSNR\,$\uparrow$ & SSIM\,$\uparrow$ & LPIPS\,$\downarrow$
        & BA-ASD\,$\downarrow$ & TVR\,$\uparrow$ & HV\,$\downarrow$ \\
\midrule
NeBLa~\cite{park2024nebla}
  & -- & -- & \checkmark & --
  & $20.79_{0.81}$ & $74.42_{0.84}$ & $0.297_{0.006}$
  & $35.67_{21.97}$ & $74.12_{3.64}$ & $3.99_{5.33}$ \\
NeBLa 320-ray
  & -- & -- & \checkmark & --
  & $20.89_{0.49}$ & $74.26_{0.78}$ & $0.286_{0.004}$
  & $35.66_{21.79}$ & $81.96_{6.70}$ & $9.39_{9.52}$ \\
\midrule
\#1. w/o MV, w/o Ref, fixed$^\dagger$ & & & & & $22.97_{0.62}$ & $78.75_{0.92}$ & $0.291_{0.007}$ & $12.45_{16.14}$ & $86.38_{3.59}$ & $3.49_{7.18}$ \\
\#2. w/o MV, w/o Ref$^\dagger$
  & & & & \checkmark
  & $23.08_{0.66}$ & $78.92_{0.99}$ & $0.289_{0.007}$ & $12.98_{17.04}$ & $\mathbf{88.16_{5.69}}$ & $2.45_{4.90}$ \\
\#3. w/o MV
  & & \checkmark & \checkmark & \checkmark
  & $22.87_{0.55}$ & $78.31_{0.74}$ & $0.287_{0.007}$ & $22.91_{18.63}$ & $85.68_{4.46}$ & $3.04_{5.64}$ \\
\#4. w/o Crs
  & \checkmark & & \checkmark & \checkmark
  & $\mathbf{23.37_{0.65}}$ & $79.20_{1.08}$ & $0.273_{0.008}$ & $7.75_{11.29}$ & $88.03_{4.40}$ & $\mathbf{1.48_{2.46}}$ \\
\#5. fixed anchors
  & \checkmark & \checkmark & \checkmark &
  & $23.26_{0.63}$ & $79.22_{1.03}$ & $0.282_{0.008}$ & $7.40_{14.68}$ & $86.70_{5.59}$ & $2.82_{4.93}$ \\
\midrule
\textbf{X-Splat (ours)}
  & \checkmark & \checkmark & \checkmark & \checkmark
  & $23.27_{0.91}$ & $\mathbf{79.62_{1.76}}$ & $\mathbf{0.259_{0.015}}$
  & $\mathbf{2.39_{0.74}}$ & $84.96_{4.51}$ & $2.96_{5.16}$ \\
\bottomrule
\end{tabular}
\end{table*}%
\indent
The contrast is starkest on canal recovery (CVR): X-Splat recovers $67.33\%$ of the MC, versus $1.14\%$ for NeBLa and $27.62\%$ for the only other competitive method, Residual CNN~\cite{henzler2018single}. Teeth recall remains on par with the strongest baselines ($84.96\%$ TVR vs.\ NeBLa $74.12\%$ and GAN~\cite{goodfellow2014generative} $70.29\%$), so the geometric gains do not sacrifice tooth coverage. X-Splat also hallucinates the least anatomy ($2.96\%$), indicating that multi-view DRR regularization suppresses false positives without hurting recall.\\
\indent
Oral-3D~\cite{song2021oral} reports strong intensity scores, but only within the reconstructed dental-arch sub-volume under masked evaluation, with geometry-aware scores relying on oracle arch geometry from ground-truth labels. Even within this region, X-Splat surpasses it on canal recall ($67.33\%$ vs.\ $48.35\%$), recovering fine structures more completely without privileged geometric access.\\
\begin{figure}[t]
\centering
\includegraphics[width=\linewidth]{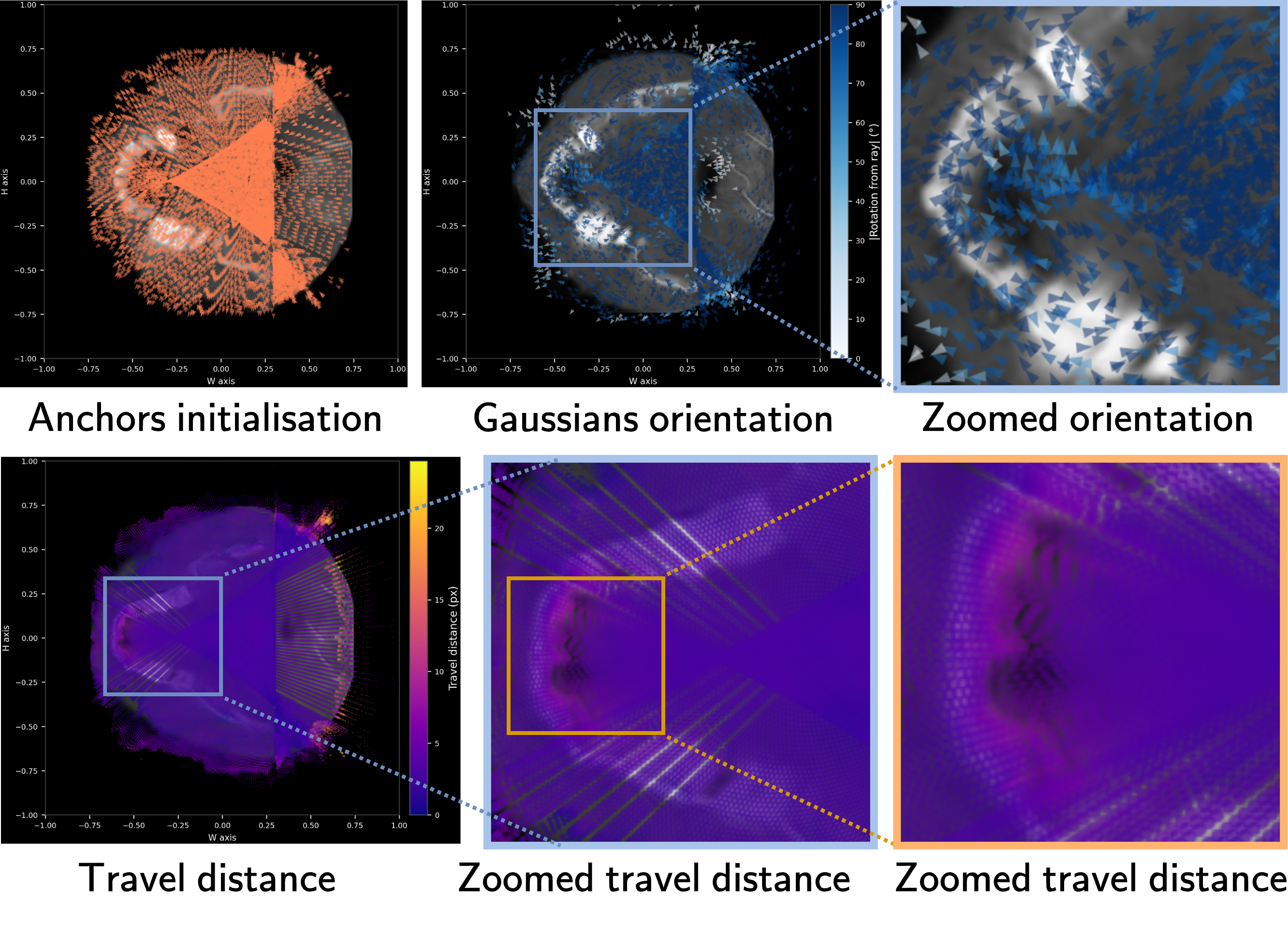}
\caption{Gaussian primitive dynamics during inference on a test case.
\textit{Top}: anchor initialization aligns splat major axes to the ray
direction (left); after training, splats rotate around the $Z$-axis (yaw
only), with the zoom revealing orientations following tooth boundaries
(colour: absolute rotation from ray direction in degrees).
\textit{Bottom}: travel distance of displaced Gaussians (stationary splats
below $1$\,px hidden; colour: displacement in pixels). Gaussians near the
dental arch migrate up to $10$\,px toward tissue boundaries, leaving voids
at their origin positions.Every tenth splat per ray shown; visualization density threshold $> 0.001$.}
\label{fig:gaussian_dynamics}
\end{figure}%
\noindent
\textbf{Qualitative results.} Figure~\ref{fig:qualitativeresults} shows generations for a representative case. X-Splat recovers the full craniofacial anatomy of the ground-truth CBCT: cortical and trabecular bone are clearly separated, the maxillary sinuses form well-defined hypodense cavities, and the inferior alveolar nerve canal appears as a continuous tubular structure through the mandible. The anisotropic Gaussians drive the sharp cortical boundaries, rendering the dense mineralized jaw shell as a bright high-attenuation rim in axial and coronal sections, which demands the precise ray-boundary localization that isotropic or implicit representations lack. Figure~\ref{fig:gaussian_dynamics} visualizes the learned orientation and
displacement of Gaussian primitives, confirming that splats actively rotate
to align with anatomical boundaries and migrate toward high-density tissue
regions during inference.\\
\indent
Competing methods recover coarser anatomy with less detail. NeBLa partially reconstructs bone but blurs the trabecular--cortical interface and misses the sinuses. Residual CNN is diffuse and over-smoothed with weak cortical definition, while the GAN introduces spurious high-frequency artifacts from optimising the maximum-intensity-projection loss. Oral-3D reconstructs only the dental-arch sub-volume, omitting the surrounding bony and soft-tissue anatomy (e.g.\ pharynx and spine).\\
\indent
The segmentation row highlights a key advantage: predicted tooth labels appear only where teeth exist in the ground truth, showing that X-Splat does not hallucinate dental structures in edentulous regions, a consequence of multi-view DRR regularization penalizing any density unsupported by projections from multiple directions.\\
\noindent
\textbf{Ablation study.} Table~\ref{tab:ablation} isolates the contribution of each component; extended ablation in Appendix~\ref{app:ablation}. Densifying NeBLa's default ray count from 256 to 320 yields only marginal TVR gains, leaves BA-ASD unchanged ($35.67$ vs.\ $35.66$\,mm), and doubles hallucinated volume, showing that the NeRF implicit field, not ray density, is the bottleneck.\\
\indent
Replacing this field with fixed Gaussian anchors and no refiner alone cuts BA-ASD by $2.9\times$ ($35.67\!\to\!12.45$\,mm) and raises PSNR to $22.97$\,dB, confirming ray-anchored anisotropic Gaussians are a far better representation for this task. Allowing ray-aligned Gaussian movement further improves TVR and lowers hallucinated volume from $3.49\%$ to $2.45\%$, so migration toward anatomy acts as an implicit geometric regularizer.\\
\indent
The refiner is only beneficial under geometric supervision. Added alone, it degrades BA-ASD from $12.98$ to $22.91$\,mm and increases hallucinations, shifting the generation burden from the Gaussian stage to learned priors. Adding $\mathcal{L}_\mathrm{mv}$ gives the single largest gain in the ablation, restoring BA-ASD to $2.39$\,mm ($9.6\times$) while suppressing hallucinations, confirming that multi-view better resolves depth ambiguity and constrains the refiner.\\
\indent
The full model attains the best BA-ASD, SSIM, and LPIPS. Its TVR ($84.96\%$) is not the highest in isolation, variant~\#2 reaches $88.16\%$, but at the cost of BA-ASD ($12.98$\,mm), exposing a recall--geometry trade-off. Overall, each component addresses a distinct failure mode: Gaussians provide geometric precision, movement adds expressivity and regularization, multi-view supervision enforces depth consistency, and coarse supervision anchors geometric fidelity in the Gaussian stage before refinement.
\section{Conclusions}

We present X-Splat, a feed-forward Gaussian Splatting framework for generating CBCT-like 3D maxillofacial volumes from a single panoramic radiograph, treating the known acquisition geometry as a generation scaffold: anisotropic Gaussian primitives anchored to the X-ray paths are displaced, rotated, and scaled to recover 3D anatomical structure. Coupled with Beer--Lambert reprojection, multi-view DRR supervision, and a lightweight residual refiner, this converts an otherwise highly underdetermined problem into a geometry-constrained volumetric inference task. 

X-Splat improves over NeRF- and GAN-based baselines in big anatomy surface accuracy, mandibular canal recovery, and hallucination suppression, suggesting that explicit Gaussian representations are well suited to dental generation, where thin cortices, root boundaries, and canal structures require localized geometric control rather than unconstrained volumetric generation.\\
\indent
This work is a step toward 3D dental generation from routine panoramic imaging. While our results are consistent across both quantitative metrics and qualitative evidence, the cohort is limited to scans with full maxillofacial field of view and paired segmentation labels.
Generalization to more diverse patient populations and imaging conditions remains to be established. Our study uses CBCT-derived simulated panoramics, which provide paired supervision and controlled evaluation but do not fully capture the variability of real panoramic systems; future work will focus on multi-site training, real-PXR adaptation, and tissue-specialized Gaussian densification. By showing that ray-anchored Gaussian primitives can recover detailed maxillofacial structure from a single projection, X-Splat opens a path toward scalable 3D dental analysis in settings where CBCT is unavailable, impractical, or unnecessary.

{
    \small
    \bibliographystyle{ieeenat_fullname}
    \bibliography{main}
}


\appendix
\maketitlesupplementary

\section{Related Work} 
\label{app:related}

\subsection{3D generation from Panoramic Radiographs}

Recovering 3D oral anatomy from a panoramic radiograph has progressed from GAN-based volumetric prediction toward physics-aware implicit representations. We organize prior work by generation scope (individual teeth vs.\ the complete oral volume) and by physics grounding.

\textbf{Tooth-level generation.} A first line recovers individual tooth shapes as meshes, point clouds, or occupancy fields rather than full CBCT intensities. X2Teeth~\cite{liang2020x2teeth} was the first CNN framework to reconstruct the occupancy of all teeth from a single PXR via localization and patch-wise generation sub-networks. Occudent~\cite{park2023occudent} used neural implicit occupancy from shape and class embeddings, PX2Tooth~\cite{ma2024px2tooth} extended this to point clouds with a root-apex prior, and DTR-Net~\cite{mei2023dtr} jointly reconstructed intensities and tooth geometry in a dual space. However, all reconstruct only tooth geometry, omitting alveolar bone, the mandibular canal, the maxillary sinus, and soft tissue, and thus cannot serve as standalone tools for implant or surgical planning.

\textbf{Volumetric oral generation.} A second line recovers tooth and bony anatomy as intensity volumes. Oral-3D~\cite{song2021oral} was the first to generate a 3D oral volume from a single PXR with a GAN~\cite{goodfellow2014generative}, but requires the dental arch curve as input, not available at inference, and resolves only a rectified box around it, excluding surrounding anatomy such as the spine. It further limits generation to dense tissues such as bone and teeth, omitting soft-tissue structures including the pharynx, maxillary sinuses, and the mandibular canal. 3DPX~\cite{li20243dpx} improved depth inference with a hybrid MLP-CNN pyramid. Neither imposes a physics-consistent constraint linking the volume to the observed projection, nor regularizes against anatomical structure, leaving the output dependent on the learned mapping's generalization.

\textbf{Physics-grounded generation.} NeBLa~\cite{park2024nebla} is the most physically principled single-view method, synthesizing a synthetic PXR intermediate via the Beer--Lambert law and a ray-tracing neural model, with real PXR translated to synthetic style at inference. While this bridges the synthetic-to-real gap, the core generation remains limited. NeBLa voxelizes implicit NeRF densities into a coarse volume and passes it to a large volumetric refiner, applying supervision only to the final refined output. Without geometric constraints on the coarse volume, depth ambiguity may remain unresolved where it matters most, and the large refiner may favor memorization of population-level anatomy over recovery of input-consistent geometry. The implicit representation further risks losing sharp root boundaries and canal cortex by construction, and imposes no geometrical prior, so globally inconsistent hallucinations may arise.

\subsection{Sparse-View Neural Volumetric Generation}

A broader body of work reconstructs volumes from sparse X-ray projections, providing the representations and physical rendering models relevant here.

\textbf{NeRF-based CT.} Adapting NeRF to X-ray replaces surface-reflectance rendering with the Beer–Lambert attenuation integral. IntraTomo~\cite{zang2021intratomo} introduced a self-supervised formulation, NeAT~\cite{ruckert2022neat} added octree-based empty-space culling, and NAF~\cite{zha2022naf} specialized CBCT with hash encodings on jaw data via TIGRE~\cite{biguri2025tigre}. SAX-NeRF~\cite{cai2024structure} added the Lineformer architecture and the X3D dataset, while geometry-aware attenuation learning~\cite{liu2024geometry} and its vessel-guided extension~\cite{liu20263d} back-project multi-view features for high-quality generation from 5--10 views. In all cases, multi-view projections remain a prerequisite for resolving the volume.

\textbf{Gaussian Splatting for X-ray and CT.} X-Gaussian~\cite{cai2024radiative} first applied 3DGS to X-ray with differentiable radiative rasterization and SfM-free initialization. $R^2$-Gaussian~\cite{zha2024r2gaussian} corrected an integration bias in X-ray 3DGS and added a differentiable voxelizer, 3DGR-CT~\cite{li20253dgr} used FBP-guided initialization, and FaCT-GS~\cite{pieta2026fact} showed GS benefits are largest in the sparse-view regime. Separately, anatomy-aware supervision improves clinical quality beyond pixel-wise losses: CARE~\cite{lin2026pixel} shows PSNR/SSIM are largely insensitive to small critical structures and adds segmentation-guided completeness penalties, but operates in a self-supervised, scene-level setting driven by multi-view reprojection and acts as a post-hoc plug-in. Together, these works establish physics-consistent GS as a competitive paradigm for sparse-view CT, offering explicit geometry, fast convergence, and direct volume extraction; yet they rely on multi-view spatial redundancy and lack dataset-level priors, so they cannot resolve the null-space of a single projection.

\textbf{Single-view generation.} Unlike per-subject sparse-view optimization, single-view generation is an ill-posed one-to-many mapping requiring strong statistical priors, motivating supervised learning on paired PXR–CBCT data. In natural images, this has been addressed by GANs~\cite{goodfellow2014generative}, encoder–decoders~\cite{henzler2018single}, multi-view architectures~\cite{choy20163d}, and cross-modal synthesis~\cite{ying2019x2ct}, and more recently by large learned priors such as Wonder3D~\cite{long2024wonder3d}, LRM~\cite{hong2024lrm}, and InstantMesh~\cite{xu2024instantmesh}. The shared insight is that a prior learned from large-scale data resolves the ambiguity a single view leaves open; for X-ray, however, that prior must capture anatomical structure and geometry rather than image-space appearance.

In summary, prior PXR-to-3D methods either recover isolated tooth geometry or full volumes without physics-consistent rendering or geometric regularization; sparse-view CT methods provide the rendering tools but require multiple projections, and their supervision has been limited to scene-level settings. Combining physics-consistent Gaussian Splatting with geometry-driven supervision in the dataset-level, single-view PXR-to-CBCT regime remains open, and is the core contribution of this work.

\section{Experimental setup}
\label{app:experimental_setup}

\subsection{Dataset}
\label{app:dataset}

\noindent\textbf{ToothFairy3.}
We use data from ToothFairy3~\cite{2026MICCAI_toothfairy3,2024IEEEACCESS,2025CVPR},
a publicly available CBCT dataset released in conjunction with the
ToothFairy3 MICCAI-ODIN 2025 challenge~\cite{2026MICCAI_toothfairy3}.
The dataset comprises three acquisition subsets that differ in scanner and
field-of-view (FoV): Set~A (417 volumes) and Set~C (52 volumes), both restricted to
the mandibular region or limited in FoV with partially cropped maxilla, acquired to capture the inferior alveolar nerve canal; and Set~B (63 volumes), acquired with a broader FoV covering the complete maxillofacial anatomy, including all upper and lower teeth from crown to root apex, the maxillary sinuses, pharynx, and surrounding bony structures. Segmentation labels follow an extended FDI World Dental Federation notation covering 77 classes: jawbones, inferior alveolar and incisive canals, maxillary sinuses, pharynx, prosthetics (bridges, crowns, implants), 32 individual teeth, and the corresponding pulp chambers (32).
 
\noindent\textbf{Subset selection.}
Single-PXR-to-CBCT reconstruction requires a full maxillofacial field of
view, as the panoramic projection integrates anatomy from both arches and
surrounding structures.  We therefore restrict training and evaluation to
Set~B and further apply two quality criteria: (i) no more than 50\% missing
dentition, ensuring sufficient tooth anatomy for meaningful geometric
evaluation, and (ii) complete crown-to-apex visibility for all present teeth
with no dental structures cropped at the volume boundary.
Applying these criteria to the 63 Set~B volumes yielded 60 eligible scans
with paired segmentation labels.  These were randomly partitioned into
training ($n{=}45$), validation ($n{=}5$), and test ($n{=}10$) sets, with
the proportion of partially edentulous patients kept comparable across all
three splits.
 
\subsubsection{Segmentation Model for Geometry-Aware Evaluation}
\label{app:seg_model}
 
To evaluate the geometric fidelity of generated volumes we use an automated
segmentation model as a structure-level evaluator: the model is applied
independently to each predicted volume and to the corresponding ground-truth
CBCT, and the resulting label maps are compared to compute the geometry-aware
metrics (BA-ASD, TVR, CVR, HV) and explained in more details is Appendix \ref{app:evaluation_metrics}.
 
We use recent method GEPAR3D~\cite{szczepanski2026morphology,szczepanski2025gepar3d} which excels in CBCT segmentation generalization~\cite{szczepanski2024let}, a
feed-forward segmentation network that unifies instance-level detection
and multi-class segmentation in a single pass, adapted here to the full
ToothFairy3 label space. The model was trained on full-maxillofacial FoV CBCT scans, the
only subset of ToothFairy3 containing complete upper and lower arch
annotations and validated on the subsets (Set~A and Set~C),
which despite their restricted field of view provide independent anatomical
coverage for the lower arch, canals, and jawbone.
 
As an evaluator, GEPAR3D is applied uniformly to all methods and to
ground truth; any segmentation inaccuracies it introduces therefore
affect all predictions consistently, preserving the validity of
relative comparisons. Crucially, any missed structures produce
pessimistic metric estimates: a tooth or canal volume that the
evaluator fails to detect lowers TVR and CVR scores for every method
equally, so reported geometry-aware improvements constitute lower
bounds on true structural recovery. This conservative behaviour is
particularly relevant for fine structures such as root apices, where
segmentation uncertainty is highest and reported scores may
underestimate true recovery.

\subsection{Implementation Details}
\label{app:implementation_details}

\subsubsection{Network Architecture}
\label{app:architecture}

\paragraph{2D U-Net encoder.}
The panoramic image $\mathbf{I}$ is processed by a 2D U-Net that produces a
dense feature map $\mathbf{F}\in\mathbb{R}^{H_p\times W_p\times 128}$ at the
input resolution.  All convolutional blocks use the double-convolution design
(two $3\!\times\!3$ convolutions per block), max-pool downsampling, bilinear
upsampling, and instance normalization.  Features at a given pixel location are
read directly at the index corresponding to each Gaussian anchor's ray-pixel
mapping.

\paragraph{Shared MLP.}
A single eight-layer MLP $f_\phi$ of width 128 is shared across all $N_g$
anchors.
For anchor $j$, the encoder feature $\mathbf{F}(\mathbf{p}_j)$ and the
sinusoidal positional encoding $\mathrm{PE}(\mathbf{p}_j)$ with $L=7$
frequency bands are each passed through a learned linear projection and summed
to form an initial fused representation $\mathbf{h}_0$, queried at a single
pixel following~\cite{mildenhall2021nerf,park2024nebla}.
Eight subsequent layers with ReLU activations process $\mathbf{h}_0$; a skip
connection at the network midpoint concatenates $\mathbf{h}_0$ back into the
hidden state before the second half of the network, following the NeRF
residual conditioning scheme~\cite{mildenhall2021nerf}.
$f_\phi$ outputs the raw displacement scalar $\hat{\delta t}_j$, log-scale
vector $\mathbf{s}_j\in\mathbb{R}^3$, unit quaternion
$\mathbf{q}_j\in\mathbb{R}^4$, and scalar density $\alpha_j$.

\paragraph{Gaussian representation.}
A total of $N_g = 6{,}553{,}600$ anisotropic 3D Gaussians are anchored along
panoramic acquisition rays ($128\text{ slices}\times256\text{ rays}\times
200\text{ samples}$, matching the ray-sample count of the SimPX renderer).
All three scale axes are initialized uniformly to $\sigma_0=0.25\;\text{px}$
and clamped throughout training to $[0.25,\,1.0]\;\text{px}$.  Movement is
restricted to the 1-D displacement $\delta t_j = \tanh(\hat{\delta
t}_j)\cdot\delta t_{\max}$ along the anchor ray within the axial ($XY$) plane,
with $\delta t_{\max}=32$ voxels; the $Z$-component is held at zero to preserve
the dense axial coverage.  Rotation is restricted to the axial plane
(rotation around the $Z$-axis only).  Density is activated with a softplus
function, and Gaussian rasterization uses the differentiable R2-Gaussian CUDA
voxelizer~\cite{zha2024r2gaussian}.

\paragraph{3D residual refiner.}
The refiner $r_\psi$ is a compact 3D U-Net with three encoder/decoder stages
and feature widths $[32, 64, 128]$, parameterized as a residual correction
$\hat{V}_f = \hat{V}_c + r_\psi(\hat{V}_c)$.  All convolutional blocks follow
the double-convolution design with strided-convolution downsampling, trilinear
upsampling, and instance normalization.  The final $3\!\times\!3\!\times\!3$
convolutional layer is zero-initialized so that $r_\psi(\hat{V}_c)=\mathbf{0}$
at the start of training, ensuring that early gradient signal flows through the
Gaussian parameters and MLP before the refiner contributes.

A full summary of architectural hyperparameters is given in
Table~\ref{tab:arch}.

\begin{table}[ht]
\centering
\small
\caption{Architecture hyperparameters.}
\label{tab:arch}
\begin{tabular}{ll}
\toprule
\textbf{Hyperparameter} & \textbf{Value} \\
\midrule
\multicolumn{2}{l}{\textit{2D U-Net encoder}} \\
\quad Output feature channels             & 128 \\
\quad Convolution blocks                  & double ($3\times3$) \\
\quad Downsampling                        & max-pool \\
\quad Upsampling                          & bilinear \\
\quad Normalization                       & instance \\
\midrule
\multicolumn{2}{l}{\textit{Shared MLP}} \\
\quad Depth (layers)                      & 8 \\
\quad Width (channels)                    & 128 \\
\quad Positional encoding frequencies $L$ & 7 \\
\quad Feature query radius                & single pixel \\
\midrule
\multicolumn{2}{l}{\textit{Gaussian representation}} \\
\quad Total anchors $N_g$                 & 6{,}553{,}600 \\
\quad Scale type                          & anisotropic (3 axes) \\
\quad Scale initialization $\sigma_0$     & 0.25 px \\
\quad Scale clamp range                   & $[0.25,\;1.0]$ px \\
\quad Movement mode                       & 1-D along ray ($XY$) \\
\quad Max displacement $\delta t_{\max}$  & 32 vox \\
\quad Rotation mode                       & axial ($Z$-axis only) \\
\quad Density activation                  & softplus \\
\quad Voxelizer                           & R2-Gaussian ~\cite{zha2024r2gaussian} \\
\midrule
\multicolumn{2}{l}{\textit{3D residual refiner}} \\
\quad Stages                              & 3 \\
\quad Feature widths                      & $[32,\;64,\;128]$ \\
\quad Convolution blocks                  & double ($3\times3\times3$) \\
\quad Downsampling                        & strided convolution \\
\quad Upsampling                          & trilinear \\
\quad Normalization                       & instance \\
\quad Final convolution kernel            & $3\times3\times3$ \\
\quad Final convolution initialization    & zero \\
\midrule
\multicolumn{2}{l}{\textit{Volume}} \\
\quad Output shape $(X\!\times\!Y\!\times\!Z)$   & $256\times256\times128$ \\
\bottomrule
\end{tabular}
\end{table}

\subsubsection{SimPX Ray Geometry and Rendering}
\label{app:geometry}

We adopt the SimPX focal-trough geometry from NeBLa~\cite{park2024nebla}
without modification.  Table~\ref{tab:geo} lists all rendering constants.
The $K=200$ uniformly spaced samples per ray define the anchor grid together
with the $128\times256$ panoramic image dimensions, yielding $N_g=6{,}553{,}600$
anchor positions. The Beer-Lambert rendering constants ($I_0$, $\beta$,
$p_{\max}$) are inherited from the NeBLa calibration and held fixed throughout
training. Synthetic PXR--CBCT pairs are registered at no acquisition cost. Matched real pairs are clinically rare, and verifying sub-voxel registration between them is non-trivial. Synthetic training scales to any CBCT dataset and provides exact ground-truth supervision under a physically calibrated forward model.
\begin{table}[ht]
\centering
\caption{SimPX ray geometry and rendering parameters.}
\label{tab:geo}
\begin{tabular}{ll}
\toprule
\textbf{Parameter} & \textbf{Value} \\
\midrule
Coordinate system          & LPS \\
Rotation centres $n_c$     & 21 \\
Samples per ray $K$        & 200 \\
Ray-length multiplier      & 0.5 \\
Angular step (molar)       & 0.528 rad \\
Angular step (default)     & 0.600 rad \\
Angular step (front)       & 1.509 rad \\
Attenuation step $\Delta s$ (ray\_delta) & 1.35 \\
Render sample mode         & nearest-neighbour \\
$I_0$                      & 1.0 \\
$\beta$                    & $7.5\times10^{-7}$ \\
$p_{\max}$                 & 0.25 \\
HU scaling method          & direct scaling \\
\bottomrule
\end{tabular}
\end{table}

\subsection{Training Details}
\label{app:training}

\paragraph{Dataset and preprocessing.}
We use a subset of the publicly available ToothFairy3
dataset~\cite{2026MICCAI_toothfairy3,2024IEEEACCESS,2025CVPR}, retaining scans
with no more than 50\% missing dentition and a full maxillofacial field of
view, where every visible tooth is fully captured from crown to root apex
without cropping.  Applying these criteria yielded 60 eligible CBCT volumes
with paired segmentation labels, partitioned into training ($n{=}45$),
validation ($n{=}5$), and test ($n{=}10$) sets while maintaining a comparable
proportion of partially edentulous patients across all splits.

All volumes are resampled isotropically from their native resolution of $0.3\;\text{mm\,px}^{-1}$ to $0.65\;\text{mm\,px}^{-1}$,
yielding a canonical grid of $256\!\times\!256\!\times\!128$ voxels that
covers the full maxillofacial field of view within GPU memory.
Intensity volumes are resampled with trilinear interpolation.
Segmentation label maps are resampled via one-hot interpolation: each class
is first binarized into a separate channel, the resulting stack of binary
maps is interpolated linearly, and the final label is recovered by
$\operatorname{argmax}$ across channels.
Compared to nearest-neighbor resampling, this preserves the boundaries of
thin structures (root apices, mandibular canal) that
could otherwise be corrupted by the coarser target grid.
 
X-Splat is trained exclusively on synthetic Synthetic PXR--CBCT pairs: panoramic
projections are synthesized from each ground-truth CBCT using
the Beer-Lambert renderer and the scanning
geometry, yielding perfectly registered
input--output pairs at no acquisition cost. SimPXRs intensities are normalized by a single global physics constant
$p_{\max}=0.25$, the divisor already embedded in the Beer-Lambert
renderer, with the zero lower bound (background
air) left unchanged.
Unlike per-image or per-dataset-range rescaling, dividing by a fixed constant
preserves the physics-consistent correspondence between CBCT HU values and
Synthetic PXR pixel intensities: the same tissue density must map to the same
projected attenuation regardless of the patient, which any data-driven
normalization would destroy.
The value $p_{\max}=0.25$ is physically grounded: with the calibrated
Beer-Lambert parameter $\beta=7.5\!\times\!10^{-7}$ and $K=200$ ray samples,
the integrated log-attenuation across the full dental anatomy does not exceed
$p_{\max}$ in any scan in the dataset, so no clipping of Synthetic PXR values occurs.
CBCT intensities are clipped to $[\mathrm{HU}_{\min},\mathrm{HU}_{\max}]
=[0,\,4000]$ (spanning air through dense enamel) and rescaled linearly to
$[0,\,1]$.  The full dataset is pre-cached in memory at the start of
training (\texttt{cache\_rate}\,=\,1.0).

\paragraph{Multi-view DRR supervision.}
During training $N=31$ azimuthal DRR projections are rendered at $\Delta\theta=7.5^{\circ}$
spacing over $\pm112.5^{\circ}$ ($225^{\circ}$ total arc, including the $0^{\circ}$
centre view).  Perceptual supervision is applied to the fine volume $\hat{V}_f$
at the eight symmetric angles
$\theta\in\{{\pm}37.5^{\circ},{\pm}52.5^{\circ},{\pm}67.5^{\circ},{\pm}82.5^{\circ}\}$
(DRR indices 4, 6, 8, 12, 18, 22, 24, 26 in the ordered stack).

\paragraph{Loss weights.}
All losses are computed independently for the coarse volume $\hat{V}_c$ and the
refined volume $\hat{V}_f$ and summed.  Perceptual losses are applied to
$\hat{V}_f$ only.  Table~\ref{tab:loss} lists all weights; note that MIP-axis
weights are uniform ($x:y:z=1:1:1$).

\begin{table}[h]
\centering
\footnotesize
\setlength{\tabcolsep}{5pt}
\caption{Loss weights for each training stage.
         ``---'' denotes a term not applied to that stage.}
\label{tab:loss}
\begin{tabular}{llrr}
\toprule
\textbf{Loss} & \textbf{Symbol} & \textbf{Fine $\hat{V}_f$} & \textbf{Coarse $\hat{V}_c$} \\
\midrule
Volumetric MSE                  & $\lambda_{\mathrm{rec}}$          & 10    & 5   \\
Beer-Lambert reprojection       & $\lambda_{\mathrm{proj}}$         & 50    & 50  \\
MIP projection (3 axes)         & $\lambda_{\mathrm{mip}}$          & 10    & 5   \\
Multi-view DRR                  & $\lambda_{\mathrm{mv}}$           & 150   & 50  \\
Perceptual (Beer-Lambert)       & $\lambda_{\mathrm{perc,proj}}$    & 0.05  & --- \\
Perceptual (MIP)                & $\lambda_{\mathrm{perc,mip}}$     & 0.01  & --- \\
Perceptual (multi-view DRR)    & $\lambda_{\mathrm{perc,mv}}$      & 0.02  & --- \\
\bottomrule
\end{tabular}
\end{table}

Weights were calibrated once on a held-out validation case after training had
stabilized at approximately 100 epochs and were then fixed for all reported
experiments.

\paragraph{Optimizer and learning-rate schedule.}
All parameters are trained with AdamW.  Per-parameter-group overrides are used
to account for the distinct roles of encoder, MLP, and refiner
(Table~\ref{tab:optim}); parameters not assigned to a named group fall back to
the global settings.The MLP receives a modestly elevated learning rate~\cite{zha2024r2gaussian}, reflecting its primary responsibility for solving the 3D geometric lifting problem from the 2D conditioning signal. A cosine annealing schedule reduces every group's learning rate from its initial value to $\eta_{\min}=10^{-5}$ over 500 epochs with no restarts.

\begin{table}[h]
\centering
\caption{AdamW hyperparameters per parameter group.
         Global $\beta_1 = 0.9$, $\beta_2 = 0.999$ (PyTorch defaults).}
\label{tab:optim}
\begin{tabular}{lrr}
\toprule
\textbf{Parameter group} & \textbf{Learning rate} & \textbf{Weight decay} \\
\midrule
Global (default)      & $1\times10^{-3}$   & $1\times10^{-6}$ \\
\quad Encoder         & $1\times10^{-3}$   & $1\times10^{-4}$ \\
\quad MLP             & $1.2\times10^{-3}$ & $1\times10^{-6}$ \\
\quad Refiner         & $1\times10^{-3}$   & $1\times10^{-4}$ \\
\midrule
Scheduler             & \multicolumn{2}{l}{cosine, $\eta_{\min}=10^{-5}$} \\
\bottomrule
\end{tabular}
\end{table}

\paragraph{Training schedule and initialization.}
Models are trained for 500 epochs with batch size 1 and random seed 42.
The residual refiner is active from epoch 0.
X-Splat is warm-started from a prior checkpoint that does not include the
refiner; on loading, any mismatched or absent parameter keys are discarded and
newly added parameters (refiner weights) are initialized from scratch.
The best checkpoint is selected based on loss value, which consists of raw unscaled value of projection (MIP) and MSE 3D losses. 

\paragraph{Hardware and software.}
All experiments are run on a single NVIDIA GPU A100 80GB with PyTorch mixed
precision in \texttt{bfloat16}. cuDNN benchmark mode is enabled.
The R2-Gaussian voxelizer is compiled as a custom CUDA extension.
No gradient clipping is applied.

\subsection{Evaluation metrics}
\label{app:evaluation_metrics}

\begin{table*}[t]
\centering
\begin{threeparttable}
\caption{Overlap-based comparison supplementing quantitative results table.
\textbf{Thr.\,DSC} (\%\,$\uparrow$): volumetric overlap at intensity
threshold $0.4$ (${\approx}600$\,HU), as used by prior
works~\cite{park2024nebla,song2021oral}; evaluates dense mineralized
structures only.
\textbf{Seg.\,DSC} (\%\,$\uparrow$): overlap against binarized
ground-truth big anatomy segmentation labels, removing threshold sensitivity. \textbf{BA-ASD} (mm\,$\downarrow$) — surface
distance over five major maxillofacial structures, with a diagonal
penalty for absent structures~\cite{bolelli2025segmenting};
\textbf{TVR} (\%\,$\uparrow$) — fraction of ground-truth tooth volume
reconstructed, averaged per present tooth class;
\textbf{CVR} (\%\,$\uparrow$) — fraction of mandibular canal volume
reconstructed;
\textbf{HV} (\%\,$\downarrow$) — falsely predicted tooth volume as a
fraction of ground-truth teeth volume.
Best per column in \textbf{bold}; second best \underline{underlined}.
$^\dagger$Intensity metrics computed over arch sub-volume only;
$^\ddagger$uses dental arch geometry from GT segmentation labels
(oracle geometry).}
\label{tab:supp_dice}
\begin{tabular}{lcc cccc}
\toprule
& \multicolumn{2}{c}{Overlap-based} & \multicolumn{4}{c}{Geometry-aware} \\
\cmidrule(r){2-3}\cmidrule(l){4-7}
Method
  & Thr.\,DSC\,$\uparrow$
  & Seg.\,DSC\,$\uparrow$
  & BA-ASD\,$\downarrow$
  & TVR\,$\uparrow$
  & CVR\,$\uparrow$
  & HV\,$\downarrow$ \\
\midrule
X2CT-GAN~\cite{ying2019x2ct}
  & $16.11_{13.22}$  & $18.57_{7.77}$
  & $58.60_{15.98}$ & $25.95_{23.86}$ & $1.88_{2.26}$   & $4.38_{8.00}$   \\
R2N2~\cite{choy20163d}
  &  $33.12_{7.64}$  & $34.57_{10.61}$
  & $28.98_{19.01}$ & $36.08_{7.64}$  & $0.01_{0.02}$   & $19.27_{16.46}$ \\
Residual CNN~\cite{henzler2018single}
  & $38.78_{9.16}$  & $41.09_{13.24}$
  & $13.06_{12.29}$ & $66.80_{9.58}$  & $\underline{27.62_{21.28}}$ & $14.95_{18.25}$ \\
GAN~\cite{goodfellow2020generative}
  & $43.08_{5.89}$  & $\underline{44.90_{6.75}}$
  & $\underline{10.60_{16.59}}$ & $70.29_{10.50}$ & $9.11_{7.70}$ & $14.36_{15.04}$ \\
NeBLa~\cite{park2024nebla}
  & $\underline{43.41_{7.26}}$  & $36.08_{11.71}$
  & $35.67_{21.97}$ & $\underline{74.12_{3.64}}$ & $1.14_{1.27}$ & $\underline{3.99_{5.33}}$ \\
\cmidrule(l){1-1}\cmidrule(l){2-3}\cmidrule(l){4-7}
Oral-3D~\cite{song2021oral}
  & $66.61_{5.89}$$^{\dagger\ddagger}$ & $38.54_{8.96}$$^{\dagger\ddagger}$
  & $47.55_{18.27}$$^\ddagger$ & $76.58_{10.79}$$^\ddagger$
  & $48.35_{22.73}$$^\ddagger$ & $1.07_{1.47}$$^\ddagger$ \\
\midrule
\textbf{X-Splat (Ours)}
  & $\mathbf{45.76_{8.59}}$  & $\mathbf{53.68_{11.39}}$
  & $\mathbf{2.39_{0.74}}$ & $\mathbf{84.96_{4.51}}$
  & $\mathbf{67.33_{19.83}}$ & $\mathbf{2.96_{5.16}}$ \\
\bottomrule
\end{tabular}
\end{threeparttable}
\end{table*}

Prior reconstruction methods~\cite{park2024nebla, song2021oral} evaluate
volumetric overlap using an intensity threshold to binarize predictions,
a reasonable proxy when segmentation labels are unavailable. However,
this approach carries two limitations that are particularly pronounced
in single-view CBCT generation. First, the threshold must be chosen to
separate tissue from background in predictions whose intensity
distributions vary across methods, introducing a systematic bias that
favors methods whose output statistics happen to align with the chosen
value. Second, depth along the projection axis is inherently ambiguous
in single-view reconstruction, so a geometrically plausible prediction
that recovers all anatomical structures at a slightly shifted position
is penalized as severely as one that misses them entirely. Critically,
a model that generates anatomy close to the population mean will score
highly on threshold DSC regardless of whether it reflects the input
radiograph, rewarding hallucination of average anatomy over
patient-specific reconstruction. The availability of dense segmentation
labels in ToothFairy3 enables us to move beyond this proxy. We
therefore prioritize surface-distance metrics expressed in millimeters,
which assess geometric fidelity directly in physical space and are
largely immune to this ambiguity, alongside volume-recall and
hallucination measures that capture clinically relevant completeness
and specificity.

\textbf{Big Anatomy ASD} (BA-ASD, mm\,$\downarrow$) reports the
symmetric Average Surface Distance over the five major craniofacial
structures $\mathcal{K}$ (mandible, maxilla, bilateral maxillary
sinuses, pharynx)~\cite{bolelli2026multi}:
\begin{equation}
\begin{split}
\text{BA-ASD} = \frac{1}{|\mathcal{K}|}\sum_{k \in \mathcal{K}}
    \frac{1}{2}\Bigl(
        &\frac{1}{|S_k|}\sum_{p \in S_k} \min_{q \in \hat{S}_k}
        \|p-q\|_2 \\
        +\,&\frac{1}{|\hat{S}_k|}\sum_{q \in \hat{S}_k} \min_{p \in S_k}
        \|p-q\|_2
    \Bigr),
\end{split}
\end{equation}
where $S_k$ and $\hat{S}_k$ are the ground-truth and predicted surfaces
of structure $k$. When a structure is absent from the prediction, a
finite penalty equal to half the volume diagonal is assigned following
\cite{bolelli2026multi}, ensuring all cases contribute a comparable
value without requiring a binary presence threshold.
\textbf{Teeth Volume Recall} (TVR, \%\,$\uparrow$) measures, for each
tooth class present in the ground truth, the fraction of its expected
tissue volume reconstructed by the model, averaged over all present
classes; a model that reconstructs only a subset of teeth scores
accordingly low, preventing partial reconstructors from appearing
competitive. \textbf{Canal Volume Recall} (CVR, \%\,$\uparrow$) applies
the same measure to the inferior alveolar nerve canal (mandibular nerve canal), a thin tubular
structure that is clinically critical yet difficult to recover.
\textbf{Hallucinated Volume} (HV, \%\,$\downarrow$) measures the total
volume of falsely predicted tooth structures for tooth classes absent
from the ground truth, expressed as a fraction of the ground-truth teeth
volume, directly quantifying the tendency to fabricate anatomy.
Intensity-based metrics (PSNR\,$\uparrow$, SSIM\,$\uparrow$,
LPIPS\,$\downarrow$) are retained as global fidelity measures and to
enable direct comparison with prior work that does not report
geometry-aware evaluation.\\
\indent
While Section~\ref{app:evaluation_metrics} argues that overlap-based
metrics are unreliable discriminators of reconstruction quality in this
task, we report them here for completeness and to enable direct
comparison with prior work~\cite{park2024nebla,song2021oral} that reported this criterion. Table~\ref{tab:supp_dice} reports threshold and segmentation Dice Scores alongside the geometry-aware metrics.
Thr.\,DSC is computed at $0.4$ (${\approx}600$\,HU on the normalized
$[-1000,3000]$\,HU scale), which lies above soft tissue and evaluates
dense mineralized structures only, excluding the mandibular canal,
pharynx, and sinuses that X-Splat explicitly reconstructs.
Seg.\,DSC is computed against binarized ground-truth \emph{big anatomy}
segmentation labels, directly measuring whether predicted tissue
coincides with actual anatomy rather than any high-attenuation
region. X-Splat achieves the best score on both metrics among
oracle-free full-volume methods ($45.76$ and $53.68$ respectively),
and is the only method for which Seg.\,DSC exceeds Thr.\,DSC, an
increase of $7.92$ points indicating that its predictions concentrate
high-attenuation tissue in anatomically consistent area. The contrasting
behavior of NeBLa is instructive: its Thr.\,DSC of $43.41$ is
competitive, but Seg.\,DSC drops to $36.08$, a fall of $7.33$ points.
This reversal indicates that NeBLa generates plausible high-attenuation
intensities whose spatial distribution may approximate the population mean
rather than the specific patient, aligning well with a fixed threshold
but poorly with actual labels. The insensitivity of Thr.\,DSC to geometric accuracy is further illustrated by Residual CNN and NeBLa: Residual CNN achieves a
$2.7\times$ lower BA-ASD than NeBLa ($13.06$ vs.\ $35.67$\,mm),
indicating substantially more accurate anatomical placement, yet scores
$4.63$ points \emph{lower} on Thr.\,DSC ($38.78$ vs.\ $43.41$),
demonstrating that a method generating geometrically correct anatomy at
a slightly shifted position is penalized more than one producing
mean-anatomy at the right average location. Oral-3D
exhibits the same pattern despite oracle arch geometry and evaluation
restricted to the arch sub-volume: Thr.\,DSC of $66.61$ collapses to
$38.54$ on Seg.\,DSC, suggesting that even within its privileged
reconstruction region, predicted tissue does not reliably coincide with
ground-truth dental anatomy. The divergence between Thr.\,DSC and
Seg.\,DSC across methods confirms that a fixed intensity threshold is
questionably reliable proxy for geometrical and anatomical accuracy in this task.\\
\textbf{Comparison with state-of-the-art.}
\label{app:sota}
We evaluate X-Splat against six representative single-view 3D generation methods. \textbf{NeBLa}~\cite{park2024nebla} is the current state-of-the-art for CBCT generation from a single panoramic radiograph, combining the Beer-Lambert driven panoramic ray geometry with a NeRF-style implicit field supervision; X-Splat builds directly upon its ray geometry and projection model. \textbf{Oral-3D}~\cite{song2021oral} is a GAN-based method that reconstructs 3D oral structures from a single panoramic image by learning a flattened volumetric representation that is then geometrically deformed and interpolated back into 3D space using the dental arch curve as a spatial deformation operator. In our implementation the arch is derived from ground-truth tooth label centroids and jaw bone skeleton, constituting oracle geometric access normally not available at inference time. \textbf{R2N2}~\cite{choy20163d} reconstructs 3D shapes from images via a recurrent encoder; following~\cite{park2024nebla}, the recurrent connection is removed for the single-image setting. \textbf{Residual CNN}~\cite{henzler2018single} employs an encoder-decoder network for volumetric generation. \textbf{GAN}~\cite{goodfellow2020generative} uses the Residual CNN as generator paired with an adversarial discriminator. \textbf{X2CT-GAN}~\cite{ying2019x2ct} reconstructs CT volumes from biplanar X-rays, adapted here to the single-view setting. Prior panoramic generation methods, including NeBLa and Oral-3D, focus primarily on hard dental tissues such as teeth and jaw bone. We train and evaluate all methods on the full maxillofacial volume, including soft-tissue cavities and the mandibular canal, providing a more complete assessment of generation quality.
\begin{table*}[t!]
\centering
\caption{%
  Ablation on test set over Gaussian count~$N$ and maximum along-ray travel   distance~$\delta t_{\max}$ (pixels). Primitives are sampled uniformly, retaining every $k$-th splat per ray; for $N$=3M every second splat is retained. Shaded rows repeat the shared reference ($N$=3M, $\delta t_{\max}$=32); all configurations otherwise use the full X-Splat pipeline and loss weights. \textbf{X-Splat} uses $N$=6M, $\delta t_{\max}$=32. Metric definitions follow main ablation table.%
}
\label{tab:ablation_splats}
\setlength{\tabcolsep}{4pt}
\begin{tabular}{c c | ccc | ccc}
\toprule
\multirow{2}{*}{$N$} &
\multirow{2}{*}{$\delta t_{max}$} &
\multicolumn{3}{c|}{Intensity-based} &
\multicolumn{3}{c}{Geometry-aware} \\
\cmidrule(lr){3-5}\cmidrule(lr){6-8}
& & PSNR\,$\uparrow$ & SSIM\,$\uparrow$ & LPIPS\,$\downarrow$
  & BA-ASD\,$\downarrow$ & TVR\,$\uparrow$ & HV\,$\downarrow$ \\
\midrule
1.5M  & 32 & $23.16_{0.67}$ & $78.16_{0.98}$ & $0.291_{0.006}$ & $12.63_{16.93}$ & $87.90_{4.10}$ & $3.69_{5.70}$ \\
3M  & 32 & $23.24_{0.72}$ & $79.86_{1.09}$ & $0.268_{0.007}$ & $5.15_{9.24}$ & $88.66_{5.37}$ & $3.86_{4.38}$ \\
6M  & 32 & $23.27_{0.91}$ & $79.62_{1.76}$ & $0.259_{0.015}$
  & $2.39_{0.74}$ & $84.96_{4.51}$ & $2.96_{5.16}$ \\
\midrule
3M  & 32 & $23.24_{0.72}$ & $79.86_{1.09}$ & $0.268_{0.007}$ & $5.15_{9.24}$ & $88.66_{5.37}$ & $3.86_{4.38}$ \\
3M  & 48 & $23.05_{0.70}$ & $78.61_{1.07}$ & $0.265_{0.006}$ & $6.98_{14.73}$ & $88.07_{4.04}$ & $2.78_{3.11}$ \\
3M  & 64 & $23.17_{0.70}$ & $78.21_{1.06}$ & $0.281_{0.006}$ & $4.66_{7.36}$ & $87.99_{3.54}$ & $2.81_{4.70}$ \\
\bottomrule
\end{tabular}
\end{table*}%
\subsection{Ablation study} 
\label{app:ablation}
Table~\ref{tab:ablation_splats} ablates Gaussian count and travel budget. Increasing $N$ from 1.5M to 3M produces the sharpest geometry gain in this experiment: BA-ASD falls from $12.63$ to $5.15$\,mm, indicating that at 1.5M inter-anchor gaps are wide enough that splats cannot fill the density field without approaching the scale limit, leaving regions either absent or covered by splats whose anchor features do not correspond to that location. A further doubling to 6M halves BA-ASD again ($5.15\!\to\!2.39$\,mm) while SSIM slightly decreases ($79.86\!\to\!79.62\%$), showing that geometry-aware metrics remain sensitive to splat density beyond the point at which intensity metrics saturate. We therefore select $N$=6M for the proposed model; the 3M configuration is a practical alternative under compute constraints, recovering most intensity quality at half the primitive count. \emph{This efficiency is unavailable to implicit NeRF representations, whose sample count is fixed by the ray grid and cannot be spatially concentrated on anatomy.}\\
\indent
For travel distance, $\delta t_{\max}$=32 gives the best overall balance. As we showed, most Gaussians use only a small fraction of the permitted range (Gaussian Dynamic Figure), with productive migrations concentrated as short-range adjustments toward tissue boundaries; the budget therefore acts as a regulariser rather than an active constraint. Larger values modestly reduce hallucinated volume (HV $3.86\!\to\!2.78$ at $\delta t_{\max}$=48) but degrade both SSIM and TVR. Since actual displacement remains well below the budget, a larger $\delta t_{\max}$ compresses the used portion of the tanh output into a steeper region of the curve, increasing gradient sensitivity and likely destabilizing optimisation without expanding the range productively used in practice.\\
\indent
Scale range was not ablated independently: it is determined by the panoramic ray geometry and constitutes a design constraint rather than a tunable hyperparameter. Splats are initialized at $1\sigma$=0.25\,px and allowed to grow up to a maximum whose $3\sigma$ extent does not significantly overlap the nearest neighboring ray, directly preserving the per-ray PXR conditioning that underlies the method. Relaxing this bound to compensate for sparser coverage at lower $N$ would cause splats to extend across multiple rays, so that each Gaussian's density is no longer informed exclusively by its anchor ray feature, thus invalidating the core conditioning mechanism rather than ablating $N$ in isolation. At $N$=1.5M, some inter-ray gaps cannot be filled without splats exceeding the scale bound. Along-ray displacement can reposition splats within their ray but cannot place density between rays, leaving those regions either absent from the density field or covered by splats whose anchor features do not correspond to that location. Both failure modes are tolerated by intensity-based metrics, which average over the full volume and have low sensitivity to localized spatial errors, but are directly exposed by BA-ASD, which measures surface
accuracy and captures the sharp geometric degradation at $N$=1.5M. The appropriate remedy is denser sampling, not larger splats, which is precisely what the $N$=3M and $N$=6M configurations provide. Notably, even at $N$=1.5M the model reconstructs anatomically coherent volumes, PSNR of $23.16$\,dB and TVR of $87.90\%$ remain competitive, indicating that the ray-anchored Gaussian representation is robust to moderate reductions in primitive count, with geometric precision rather than overall generation quality being the primary casualty of sparser coverage.

\subsection{Qualitative Results} 
\begin{figure*}[h]
\centering
\includegraphics[width=\textwidth]{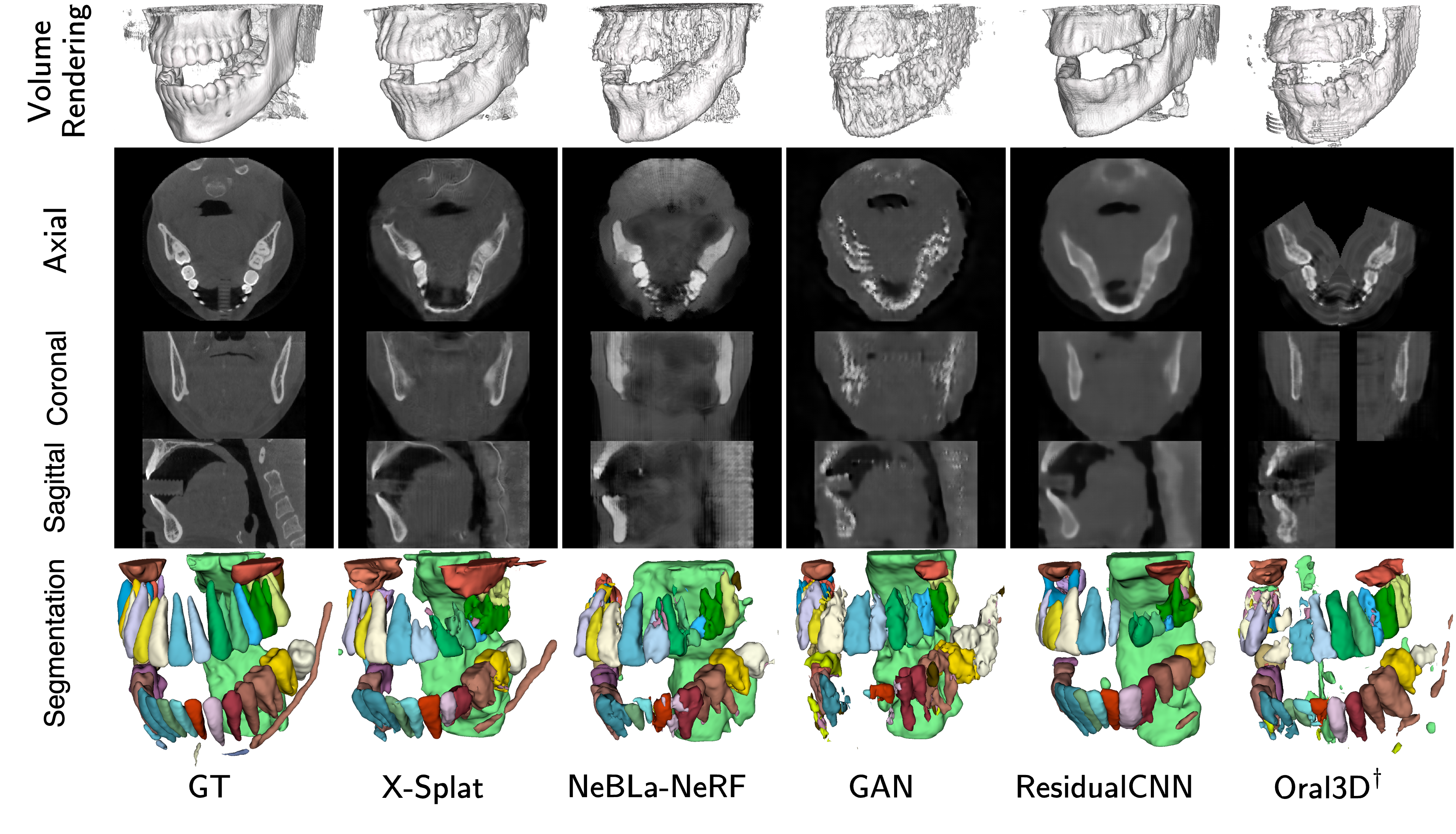}
\caption{Supplementary qualitative comparison, case ID 027. $^\dagger$Oral-3D uses oracle arch geometry.
}
\label{fig:qualitative_sup1}
\end{figure*}

\begin{figure*}[h]
\centering
\includegraphics[width=\textwidth]{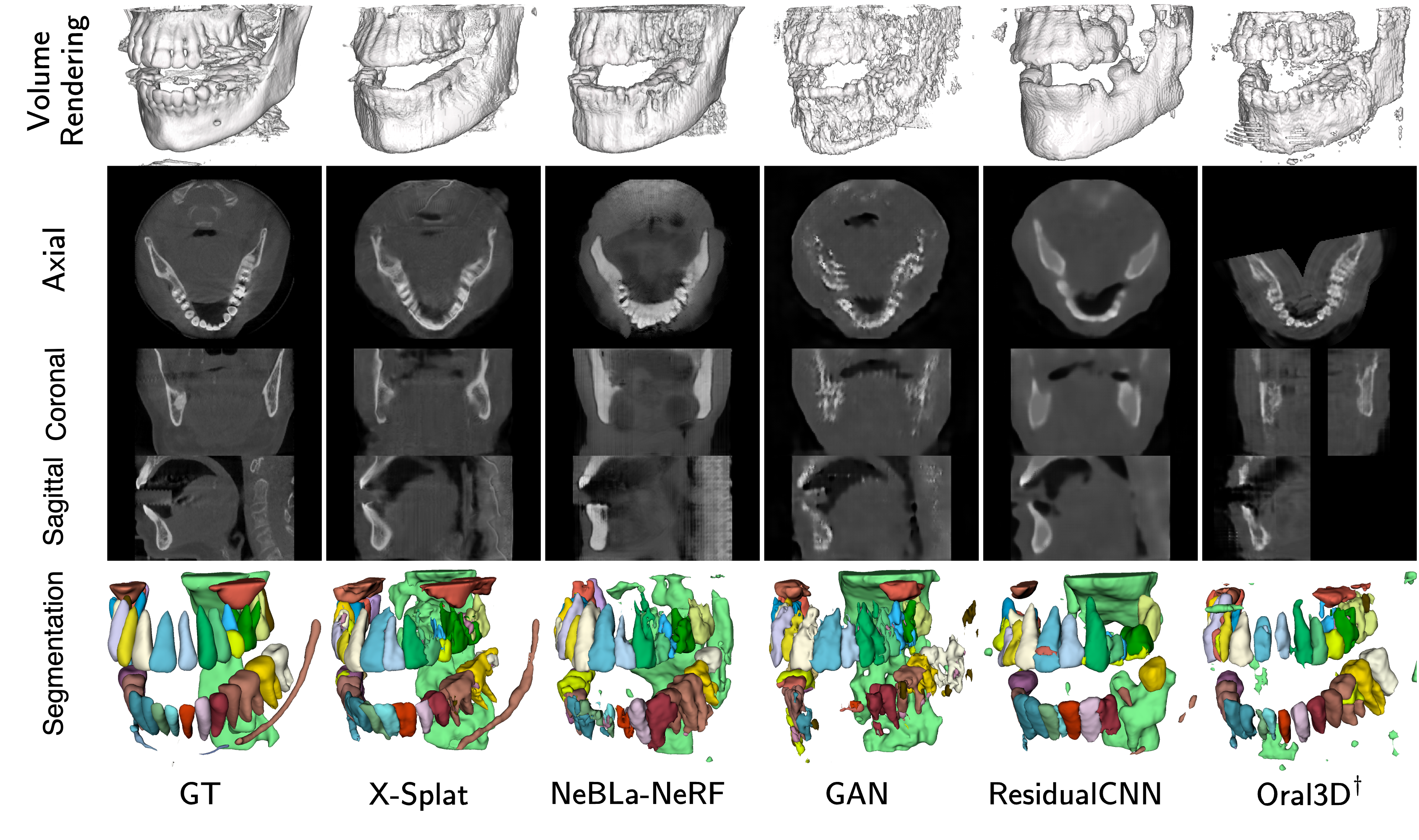}
\caption{Supplementary qualitative comparison, case ID 030. $^\dagger$Oral-3D uses oracle arch geometry.
}
\label{fig:qualitative_sup2}
\end{figure*}

Figures~\ref{fig:qualitative_sup1} and~\ref{fig:qualitative_sup2} extend the
qualitative comparison of the main paper to two additional test cases.
X-Splat is the only method to recover the mandibular canal as a
continuous tubular structure, visible as a curved hypodense channel in
the sagittal cross-section and as a distinct labelled object in the 3D
segmentation. Competing methods either omit the canal entirely or
produce disconnected fragments, consistent with the reported CVR gap. The maxillary
sinuses appear as clearly defined bilateral hypodense cavities in the
GT Sagittal view; X-Splat reproduces partially their shape and symmetry, while
NeBLa-NeRF and Residual CNN fill the sinus region with diffuse
intermediate intensity and GAN introduces artifacts that obscure the
boundary entirely. Oral-3D, despite oracle arch access, reconstructs
only the dental sub-volume and produces no sinus anatomy.\ 
Despite strong overall anatomy recovery, X-Splat struggles to precisely
reconstruct root apices, where the fine geometry may exceed the resolution
of the current splat density; larger training datasets and feed-forward adaptive
densification of Gaussian primitives near high-curvature boundaries
remain directions for future work.

\begin{figure}[h!]
\centering
\includegraphics[width=\linewidth]{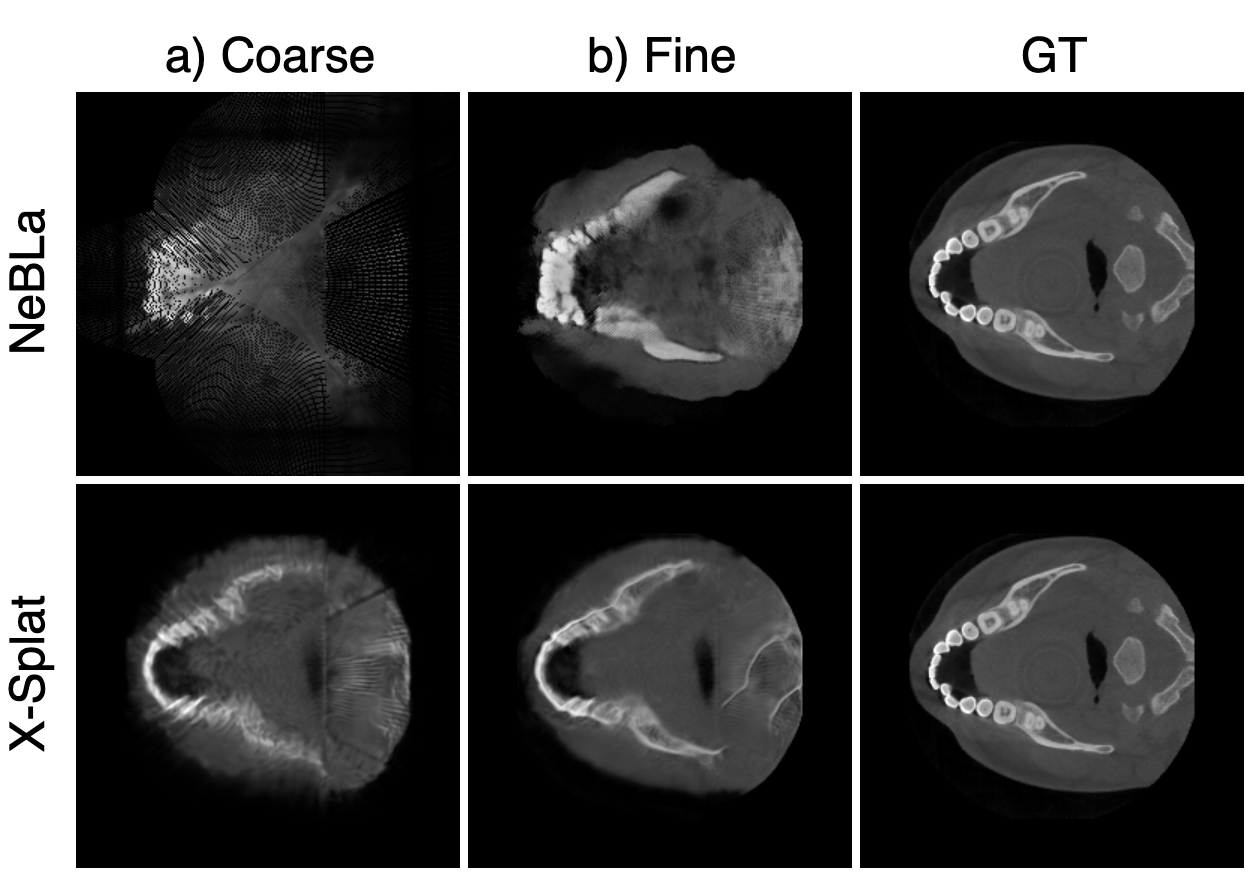}
\caption{Axial cross-sections of coarse (a) and refined (b) volumes alongside ground truth. NeBLa's coarse volume exhibits inter-ray gaps and little anatomical structure, placing geometry recovery almost entirely on the refiner. X-Splat's Gaussian voxelizer produces a continuous coarse volume with the dental arch already localized and mandibular curvature resolved; the residual refiner then performs targeted correction rather than complete reconstruction, indicating that geometric information is preserved through the pipeline rather than invented by the refiner.
}
\label{fig:coarse_fine}
\end{figure}

\end{document}